\documentclass[conference, onecolumn]{IEEEtran}
\usepackage{listings}
\usepackage{graphicx}
\usepackage{subcaption}
\usepackage[table,xcdraw]{xcolor}
\usepackage{amsmath}
\usepackage{pifont}
\usepackage{amssymb}
\definecolor{mydarkblue}{rgb}{0,0.08,0.65}
\usepackage[colorlinks=true,
    linkcolor=mydarkblue,
    citecolor=mydarkblue,
    filecolor=mydarkblue,
    urlcolor=mydarkblue]{hyperref} 
\usepackage{cleveref}
\usepackage{soul}
\usepackage[shortlabels]{enumitem}
\usepackage{url}
\usepackage{color}
\usepackage{tikz}
\usepackage{balance}
\usepackage{multirow}
\usepackage{comment}
\usepackage{wrapfig,lipsum,booktabs}
\usepackage[frozencache,cachedir=.]{minted}

\usepackage[symbol]{footmisc}
\usepackage{tablefootnote}
\usepackage{tabularx}
\usepackage{placeins}
\usepackage{algorithm}
\usepackage{algpseudocode}

\usepackage{cuted}
\usepackage{float}
\usepackage{caption}

\usepackage{multicol}

\usepackage{dblfloatfix}
\usepackage{natbib}

\renewcommand{\arraystretch}{1.2}

\definecolor{codegreen}{rgb}{0,0.6,0}
\definecolor{codegray}{rgb}{0.5,0.5,0.5}
\definecolor{codepurple}{rgb}{0.58,0,0.82}
\definecolor{backcolour}{rgb}{0.95,0.95,0.92}

\makeatletter
\def\blfootnote{\xdef\@thefnmark{}\@footnotetext}
\makeatother

\lstdefinestyle{mystyle}{
  backgroundcolor=\color{backcolour},   commentstyle=\color{codegreen},
  keywordstyle=\color{magenta},
  numberstyle=\tiny\color{codegray},
  stringstyle=\color{codepurple},
  basicstyle=\ttfamily\footnotesize,
  breakatwhitespace=false,         
  breaklines=true,                 
  captionpos=b,                    
  keepspaces=true,                 
  numbers=left,                    
  numbersep=5pt,                  
  showspaces=false,                
  showstringspaces=false,
  showtabs=false,                  
  tabsize=2,
}

\AtBeginDocument{%
  \providecommand\BibTeX{{%
    \normalfont B\kern-0.5em{\scshape i\kern-0.25em b}\kern-0.8em\TeX}}}

\renewcommand{\IEEEauthorrefmark}[1]{\textsuperscript{#1}}

\IEEEoverridecommandlockouts
\begin{document}

\title{Training Foundation Models on a Full-Stack AMD Platform: Compute, Networking, and System Design}

\newcommand{\corr}{\textsuperscript{*}}

\author{
\IEEEauthorblockN{Quentin Anthony\IEEEauthorrefmark{1}\corr, 
Yury Tokpanov\IEEEauthorrefmark{1}, 
Skyler Szot\IEEEauthorrefmark{1},
Srivatsan Rajagopal\IEEEauthorrefmark{1},
Praneeth Medepalli\IEEEauthorrefmark{1},\\
Anna Golubeva\IEEEauthorrefmark{1},
Vasu Shyam\IEEEauthorrefmark{1},
Robert Washbourne\IEEEauthorrefmark{1},
Rishi Iyer\IEEEauthorrefmark{1},
Ansh Chaurasia\IEEEauthorrefmark{1},
Tomas Figliolia\IEEEauthorrefmark{1},\\
Xiao Yang\IEEEauthorrefmark{1},
Abhinav Sarje\IEEEauthorrefmark{1},
Drew Thorstensen\IEEEauthorrefmark{2}, 
Amartey Pearson\IEEEauthorrefmark{2}, 
Zack Grossbart\IEEEauthorrefmark{2},\\
Jason van Patten\IEEEauthorrefmark{2},
Emad Barsoum\IEEEauthorrefmark{3}, 
Zhenyu Gu\IEEEauthorrefmark{3},
Yao Fu\IEEEauthorrefmark{3},
Beren Millidge\IEEEauthorrefmark{1}\corr}
\IEEEauthorblockA{\IEEEauthorrefmark{1}Zyphra}
\IEEEauthorblockA{\IEEEauthorrefmark{2}IBM}
\IEEEauthorblockA{\IEEEauthorrefmark{3}AMD}
\IEEEauthorblockA{\textsuperscript{*}Corresponding authors: \texttt{quentin@zyphra.com}, \texttt{beren@zyphra.com}}
}

\maketitle

\setcounter{page}{1}

\begin{abstract}
We report on the first large-scale mixture-of-experts (MoE) pretraining study on pure AMD hardware, utilizing both MI300X GPUs and Pollara networking. We distill practical guidance for both systems and model design. On the systems side, we deliver a comprehensive cluster and networking characterization: microbenchmarks for all core collectives (AllReduce, ReduceScatter, AllGather, Broadcast) across message sizes and GPU counts over Pollara. To our knowledge, this is the first at this scale. We further provide MI300X microbenchmarks on kernel sizing and memory bandwidth to inform model design. On the modeling side, we introduce and apply MI300X-aware transformer sizing rules for attention and MLP blocks and justify MoE widths that jointly optimize training throughput and inference latency. We describe our training stack in depth, including often-ignored utilities such as fault-tolerance and checkpoint–reshaping, as well as detailed information on our training recipe. We also provide a preview of our model architecture and base model - ZAYA1 (760M active, 8.3B total parameters MoE, available at \url{https://huggingface.co/Zyphra/ZAYA1-base}) - which will be further improved upon in forthcoming papers. ZAYA1-base achieves performance comparable to leading base models such as Qwen3-4B and Gemma3-12B at its scale and larger, and outperforms models including Llama-3-8B and OLMoE across reasoning, mathematics, and coding benchmarks. Together, these results demonstrate that the AMD hardware, network, and software stack are mature and optimized enough for competitive large-scale pretraining.
\end{abstract}

\section{Introduction}

The capabilities of a final large language model (LLM) are substantially influenced by the quality of the base model obtained during pretraining. The quality of a base model is primarily bottlenecked by the aggregate throughput of the high-performance computing (HPC) cluster and the underlying training software stack that is used to train it. Many practicalities of pretraining at scale, such as load balancing, fault tolerance, communication primitives, and low-level GPU kernels, determine whether aggregate throughput is sufficient to reach the frontier level of model performance. In this paper, we present a case study of production-scale pretraining on an AMD platform, encompassing both compute (MI300X GPU \citep{amd2024cdna3}) and interconnect (Pollara \citep{amd_pollara_2024}). In addition, we document in detail the operational challenges inherent in training a model from scratch and how established challenges map to this novel hardware setup.

For this purpose, we introduce \emph{ZAYA1-base} (\url{https://huggingface.co/Zyphra/ZAYA1-base}), a mixture-of-experts (MoE) transformer trained on an MI300X cluster with AMD Pensando Pollara networking. Architecturally, ZAYA1-base follows Zyphra's novel `MoE++' recipe that combines: (i) \emph{Compressed Convolutional Attention} (CCA) \citep{cca}, which executes attention in a compressed latent space to reduce both prefill compute and KV-cache size;\footnote{We use the term \emph{prefill} to denote the first phase of inference, where the model must first ingest the entire input sequence and generate the first token. This is equivalent to next-token prediction during pretraining. The KV-cache is a purely inference-time structure that stores the KV states of prior tokens so that they do not require recomputation for each subsequent token.} (ii) a significantly more powerful and expressive \emph{ZAYA1 router} that replaces the standard linear gate with a compact MLP with depth-aware mixing to promote expert specialization and stable load balancing; and (iii) lightweight \emph{residual scaling} that gives the layers fine-grained control over information flow through the residual stream with negligible parameter overhead (See Section \ref{sec:model}). Beyond the core model architecture, our training run used context-parallelism (CP) (Section \ref{sec:context-parallelism}) tailored to CCA, fused-optimizer and LayerNorm (LN) kernels (Section \ref{sec:kernels}), and training/IO paths tuned for the AMD software stack (Sections \ref{sec:checkpointing} and \ref{sec:fault-tolerance}). We also optimized the shapes of our model architecture for performance on AMD hardware and describe in detail how to perform this kind of model sizing to optimize training and inference on a particular hardware platform (Sections \ref{sec:model-sizing}). 

\begin{figure*}[htbp]
    \centering
    \includegraphics[width=0.65\linewidth]{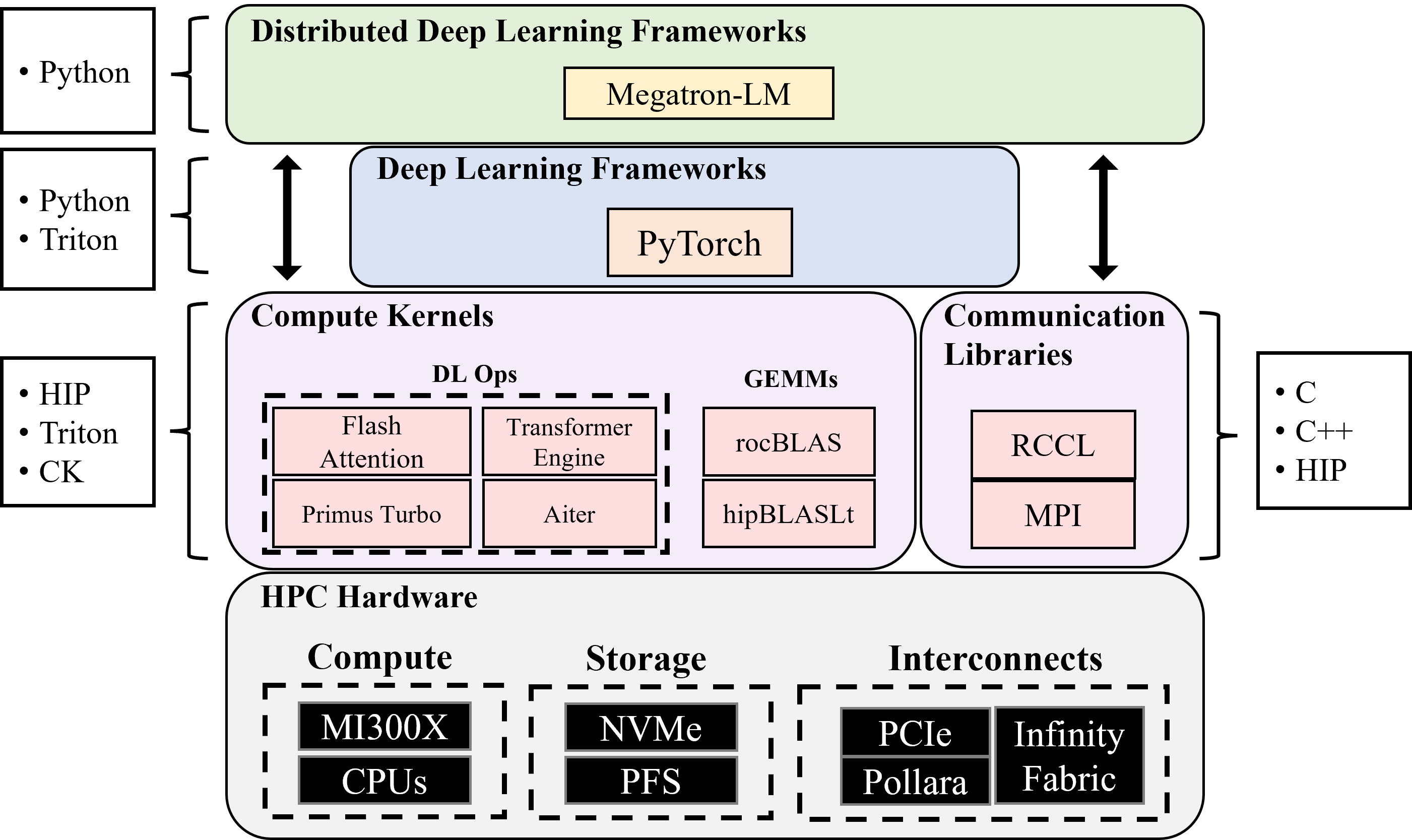}
    \caption{The AMD software stack used to train ZAYA1, along with the respective languages and component libraries that each layer is written in. The principal hardware is described in Section \ref{sec:cluster-setup}. Our core training framework is a forked internal version of Megatron-LM adapted for the AMD stack.}
    \label{fig:software-stack}
\end{figure*}

The AMD software stack is depicted in Figure~\ref{fig:software-stack}. Layers at the bottom of the stack are closer to hardware, and therefore provide more control over the compute and networking at the cost of developer productivity. As such, low layers are written using languages and libraries such as Heterogeneous-Compute Interface for Portability (HIP) \citep{amd_hip}, or Composable Kernel (CK) \citep{amd_composable_kernel}. Higher layers provide more abstraction for rapid prototyping, but hide hardware details and limit control. The predominant programming languages and libraries used to implement each layer are provided where possible.

While our results speak to model quality and efficiency, the primary contribution of this paper is a careful, measurement-driven account of how the AMD hardware, network, and software behave under LLM pretraining workloads. We quantify where time and bandwidth go in practice, which configuration choices matter most on MI300X, and how to avoid common pitfalls when porting mature NVIDIA pipelines to ROCm. To our knowledge, this study is among the first to present systematic collective-communication and memory-bandwidth microbenchmarks, end-to-end iteration breakdowns, and transformer sizing guidance at this scale on an AMD stack.

\subsection{Contributions} This paper makes the following contributions:

\begin{itemize}[leftmargin=*]
\item \textbf{End-to-end AMD pretraining case study:} We describe the engineering required to bring a large MoE pretraining job to production on MI300X with AMD networking, including kernel selection, runtime configuration, and ROCm/RCCL nuances that materially affect throughput and stability.
\item \textbf{Networking characterization at scale:} We benchmark Pollara/Pensando networking for the collectives that dominate LLM training (e.g., AllReduce, ReduceScatter, AllGather), sweeping message sizes and GPU counts. We relate these measurements to optimizer/gradient communication and to context-parallelism (CP) traffic patterns induced by CCA.
\item \textbf{Transformer sizing guidance for MI300X:} We provide sizing recommendations for attention and MLP blocks that respect MI300X compute–memory balance and NIC characteristics. We justify our small-MoE (fine-grained experts, top-1 routing) architectural choices for both training and inference efficiency.
\item \textbf{Memory-bandwidth microbenchmarks:} We present targeted HBM bandwidth tests and cross-compare against widely cited figures, clarifying what sustained bandwidth looks like for kernels that matter to LLMs rather than synthetic peaks.
\item \textbf{Cluster architecture diagrams.} We document the compute and storage node internals and the full cluster topology, highlighting bandwidth and contention points that affect scaling.
\item \textbf{Training mechanics and fault tolerance:} We detail important but under-described aspects of the training stack such as checkpointing, reshaping across parallelism regimes, and a practical fault-tolerance setup tailored to long runs (including a reshape service and accelerated checkpoint writes in PyTorch).
\item \textbf{Iteration-time breakdown:} We decompose iteration time into attention/MLP/norm compute, gradient and optimizer communication, and IO.
\item \textbf{Parallelism recipe for CCA:} We describe a CP design co-tuned with CCA that keeps activation memory and communication predictable as context length increases, and we explain how this interacts with sharded optimizer states and checkpoint reshaping.
\item \textbf{Kernel and optimizer engineering:} We include descriptions for our fused Muon optimizer kernels (including a matrix–matrix transpose kernel used in Muon), a fused RMSnorm/LN kernel, and discuss Muon optimizer settings for stable large-batch training.
\end{itemize}

\section{Cluster Setup}
\label{sec:cluster-setup}
\vspace{1ex}

\begin{figure*}[htbp]
    \centering
    \begin{minipage}[b]{0.48\textwidth}
        \centering
        \includegraphics[width=\textwidth]{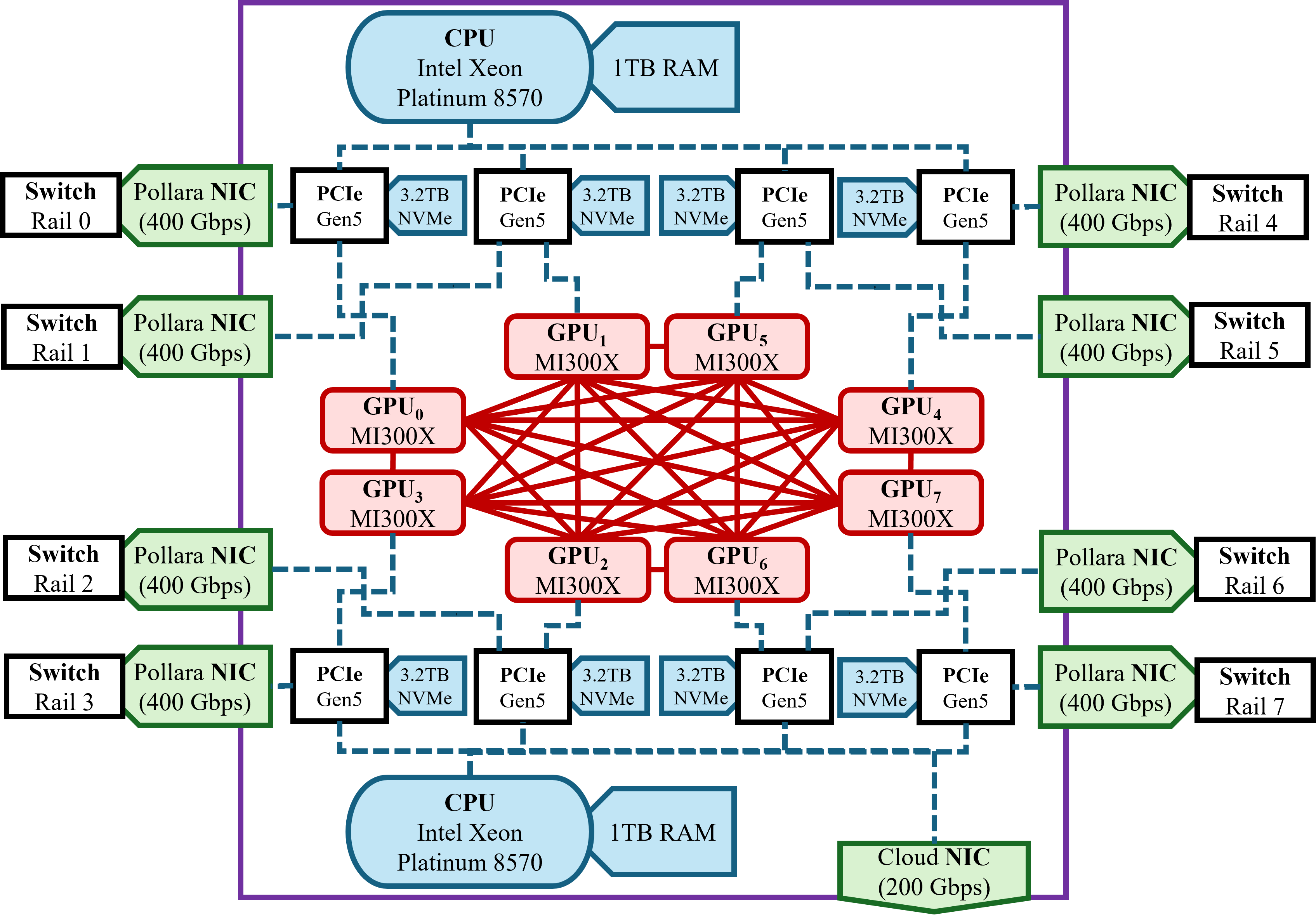}
        \caption*{(a) The architecture of a single node}
    \end{minipage}
    \hfill
    \begin{minipage}[b]{0.50\textwidth}
        \centering
        \includegraphics[width=\textwidth]{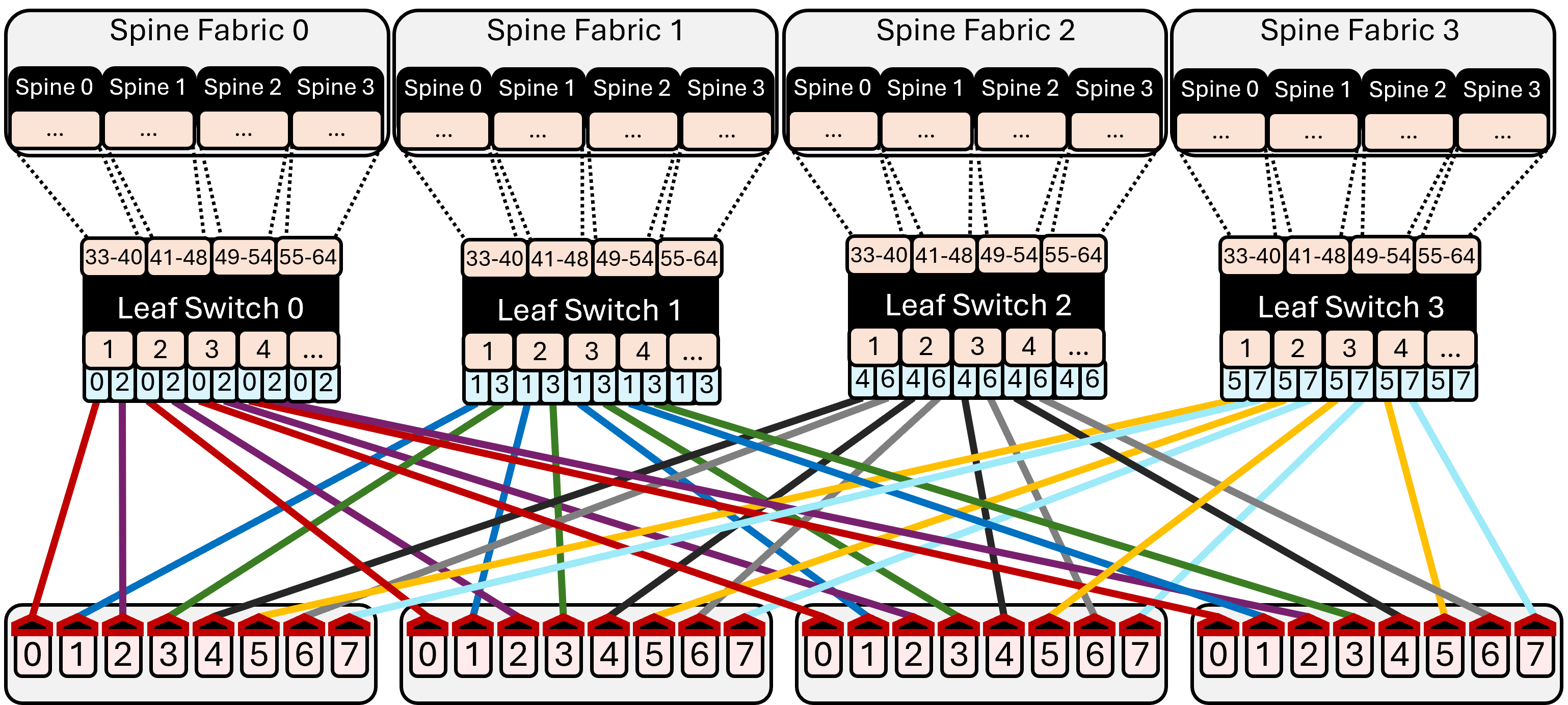}
        \caption*{(b) Cluster topology}
    \end{minipage}
    \caption{The architecture of the Zyphra pretraining cluster. Each node contains 8 MI300X GPUs interconnected with InfinityFabric. Each GPU is assigned its own Pollara 400Gbps NIC, which connects the GPU to its respective switch in a rails-only topology.  Our cluster switch topology is a two-level rails-only setup, with four physical leaf switches, and four spine switches per leaf switch. We have 4 physical leaf switches per slice. Each slice consists of 30 servers. The slices are aggregated via 4 spine switches per leaf. Each compute node is connected to the headnode and storage node via a separate interconnect to avoid contention.}
    \label{fig:cluster-arch}
\end{figure*}

To contextualize the subsequent sections, we first describe the high-level architecture.

\subsection{Node Architecture}
\label{sec:node-architecture}

Each compute node consists of eight MI300X GPUs connected via InfinityFabric, dual-socket Intel Xeon CPUs with 2\,TB of DDR5 memory, a dedicated Pollara 400~Gbps NIC per GPU, and local NVMe storage for high-throughput datasets and checkpoints (see Figure~\ref{fig:cluster-arch}). A detailed description for the compute, storage, and login nodes is provided in Table~\ref{tab:hardware_specs} and Appendix~\ref{app:cluster}.

\begin{table}[ht]
\centering
\begin{tabular}{|c|c|c|c|}
\hline
\textbf{Hardware Element} & \textbf{Compute Node} & \textbf{Storage Node} & \textbf{Login Node} \\
\hline
\textbf{GPUs} & 8 MI300X & --- & --- \\
& \citep{amd2024cdna3} & & \\
\hline
\textbf{RAM} & 2\,TB DDR5 & 256\,GB DDR5 & 80\,GB \\
& 16×128\,GB DIMMs & 16×16\,GB DIMMs & 5×16\,GB DIMMs \\
& Samsung M321RAJA0MB0-CWMNY & Samsung M321R2GA3BB6-CQKET & (virtual/QEMU) \\
& 5600 MT/s & 4800 MT/s & \\
\hline
\textbf{CPU} & 2× Intel Xeon Platinum 8570 & 2× Intel Xeon Gold 6426Y & 2× Intel Xeon \\
& 56 cores, 2 threads/core & 16 cores, 2 threads/core & (Sapphire Rapids) \\
& 1\,TB RAM per socket & 128\,GB RAM per socket & 8 cores, 2 threads/core \\
\hline
\textbf{Networking} & 8× Pollara 400 NICs (400Gbps) & 1× Pensando DSC NIC (200Gbps) & --- \\
\hline
\textbf{Storage} & 25.6\,TB & 120\,TB &  \\
& 8× NVMe drives (3.2\,TB each) & 16× NVMe drives ($\approx7.6$\,TB each) & $\approx1$\,TB \\
& Micron MTFDKCC3T2TGQ & RAID0 array (\texttt{/dev/md127}) &  \\
\hline
\end{tabular}
\caption{Hardware specifications for compute, storage, and login nodes. Each node type has separate local drives for the OS.}
\label{tab:hardware_specs}
\end{table}

\subsection{Cluster Architecture}

Instead of the classical Clos \citep{clos} topology, individual nodes are interconnected in a rails-only topology, which was popularized in \citet{meta-rails} as a way to exploit the nature of 3D-parallel deep learning training communication patterns to reduce the number of Arista 7060X6-64PE \citep{arista7060x6} switches. Specifically, our cluster switch topology is a two-level rails-only setup, with four physical leaf switches, and four spine switches per leaf switch. This design trades some path diversity and routing flexibility compared to a full Clos fabric for lower cost on physical switches and simpler wiring. In practice, this means the topology leads to a slight performance penalty compared to a Clos network unless the parallelism topology and collective communication algorithm is careful to avoid cross-rail traffic.

Our cluster employs two physically separate networks. The Pollara training fabric handles all collective communication for distributed training, while a VPC network manages dataset I/O, checkpoints, and cluster management. This separation prevents storage operations from interfering with model communication (gradients, optimizer states, etc). The VPC network connects each node via Pensando DSC SmartNIC/DPUs to a separate leaf--spine fabric, providing approximately 200 Gbps throughput to the storage node from any of the compute nodes.

\section{Hardware Characterization}
\label{sec:hardware-characterization}

To determine the optimal model architecture (including sizing and component kernels) and training parallelism strategy, we first performed an in-depth analysis of the hardware characteristics for representative deep learning (DL) workloads. This must be performed for the GPU (both compute and memory), and all available interconnects (InfinityFabric \citep{amd2024cdna3} intra-node and Pollara \citep{amd_pollara_2024} inter-node). DL-specific operations like attention, LayerNorm, and MLPs are not considered here. Instead, we seek to boil down those ops into their component set of memory accesses and General Matrix Multiplications (GEMMs), then define amenable sizes and kernels for these operations on our target training and inference hardware.

\subsection{GPU HBM Bandwidth}
\label{sec:memory-bandwidth}

MI300X GPUs have 192\,GB HBM, but the speed of writes/reads to and from HBM constitutes a core bottleneck for some intermediate operations such as the LayerNorm, activations, and even attention for shorter context lengths. Even IO-aware attention algorithms like Flash Attention \citep{dao2023flashattention2} are bottlenecked by the bandwidth of their HBM accesses until they reach higher sequence lengths. The crossover point is dependent on the particular GPU's roofline, but the crossover point from being linearly HBM-bound to quadratically compute-bound tends to fall around 4-16k sequence length, which is higher than many practitioners expect.

Choosing a fair software setup and benchmarking regime for measuring HBM bandwidth, as with any other hardware metric, is challenging. In order to test the hardware's limits, vendors and kernel engineers opt for low-level benchmarks with custom data transfer kernels such as nvbandwidth \citep{nvbandwidth} and the ROCm bandwidth tests (RBT) \citep{rocm_systems}. When conveying application results such as DL frameworks, these low-level workloads are often not representative of the available memory transfer kernels, even when under full load. However, if application benchmarks are exclusively relied upon, one is indirectly measuring both: 1) How well that application is optimized for the hardware in question, which heavily favors the incumbent, and 2) What overhead the library incurs. Note that the severity of these effects differs per hardware element and its underlying kernels. Specifically, while PyTorch communication operations and compute GEMMs are nearly a direct mapping to their underlying libraries (e.g. RCCL and rocBLAS, respectively), hardware like HBM and storage (NVMe/SSDs) must be handled more carefully.

\begin{figure*}[htbp]
    \centering
    \includegraphics[width=0.7\linewidth]{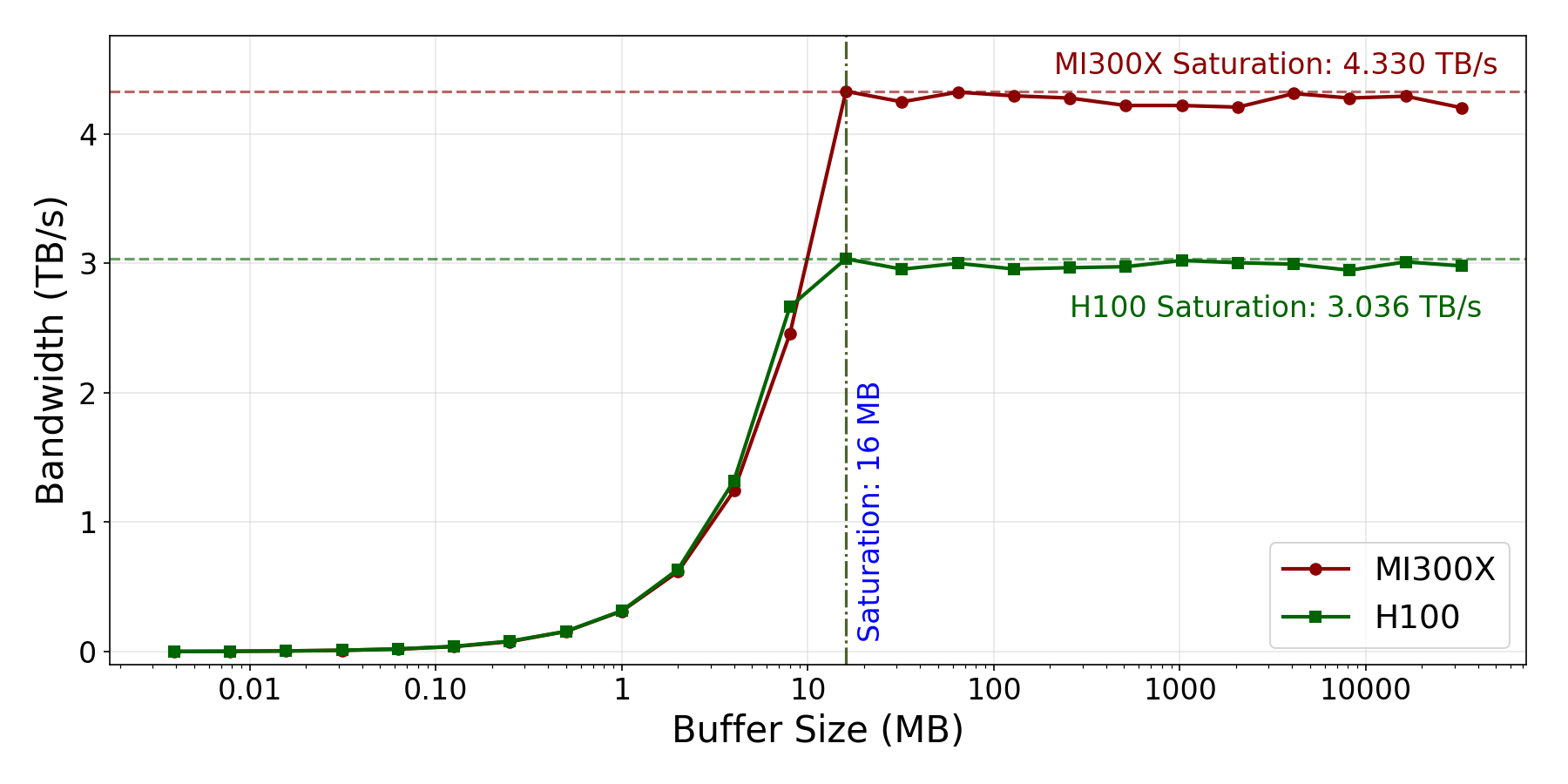}
    \caption{The achievable memory bandwidth to HBM for PyTorch using the ROCm/CUDA backends.}
    \label{fig:bandwidth}
\end{figure*}

With this in mind, we implemented a PyTorch benchmark that mirrors the memory access patterns of a benchmark such as \citep{babelstream} by carefully laying out the tensor when reading. Further, we tuned the memcpy kernel within PyTorch to better select the kernel thread block size depending on the tensor size for the MI300X backend. While not all of the memory access patterns incurred by our workloads are so clean (e.g. non-contiguous tensors and sparse access patterns), we believe these results to better capture what the HBM hardware is capable of under real training and inference workloads. The results of which are depicted in Figure~\ref{fig:bandwidth}.

\subsection{GPU Compute}
\label{sec:gpu-compute}

\begin{figure*}[htbp]
    \centering
    \begin{minipage}[b]{0.48\textwidth}
        \centering
        \includegraphics[width=\textwidth]{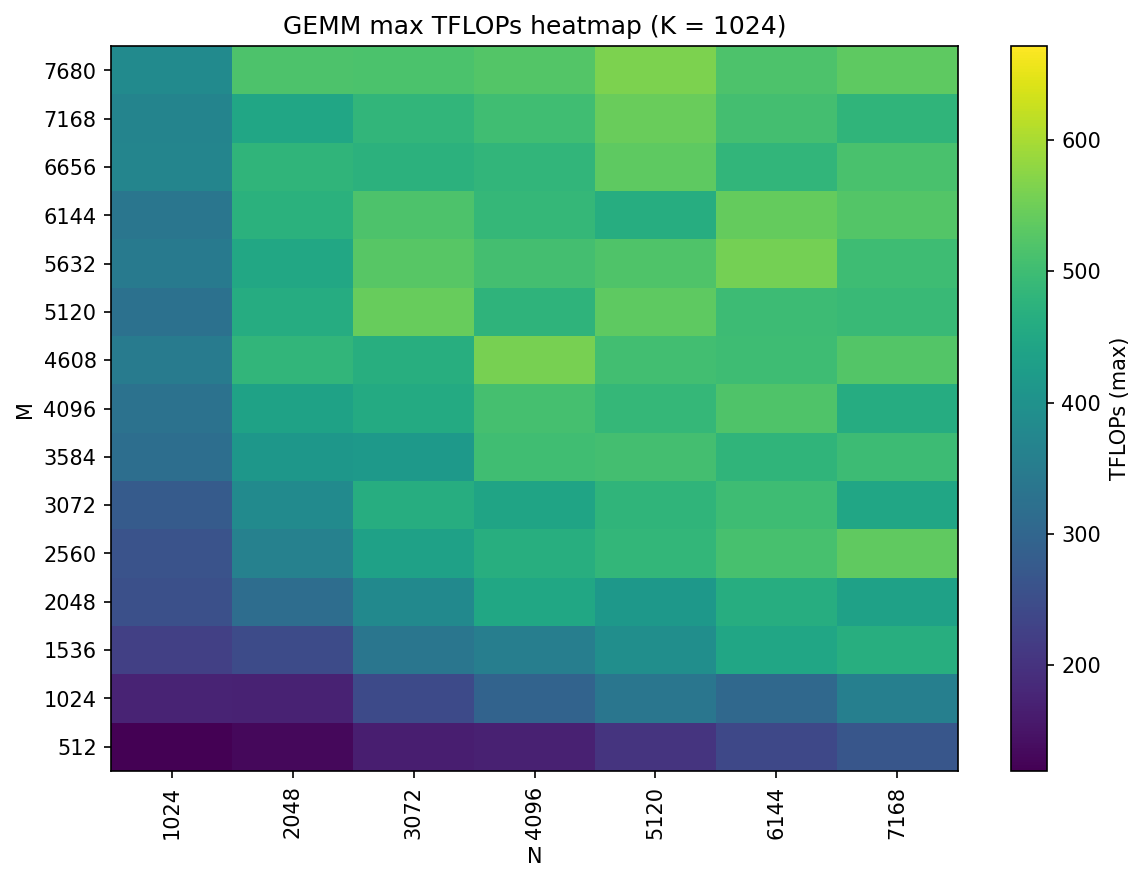}
        \caption*{(a) GEMM performance (TFLOPS/s) on MI300X with K=1024. Small $K$ limits throughput; large M and N are needed to approach peak performance. Performance varies significantly even at large problem sizes.}
    \end{minipage}
    \hfill
    \begin{minipage}[b]{0.50\textwidth}
        \centering
        \includegraphics[width=\textwidth]{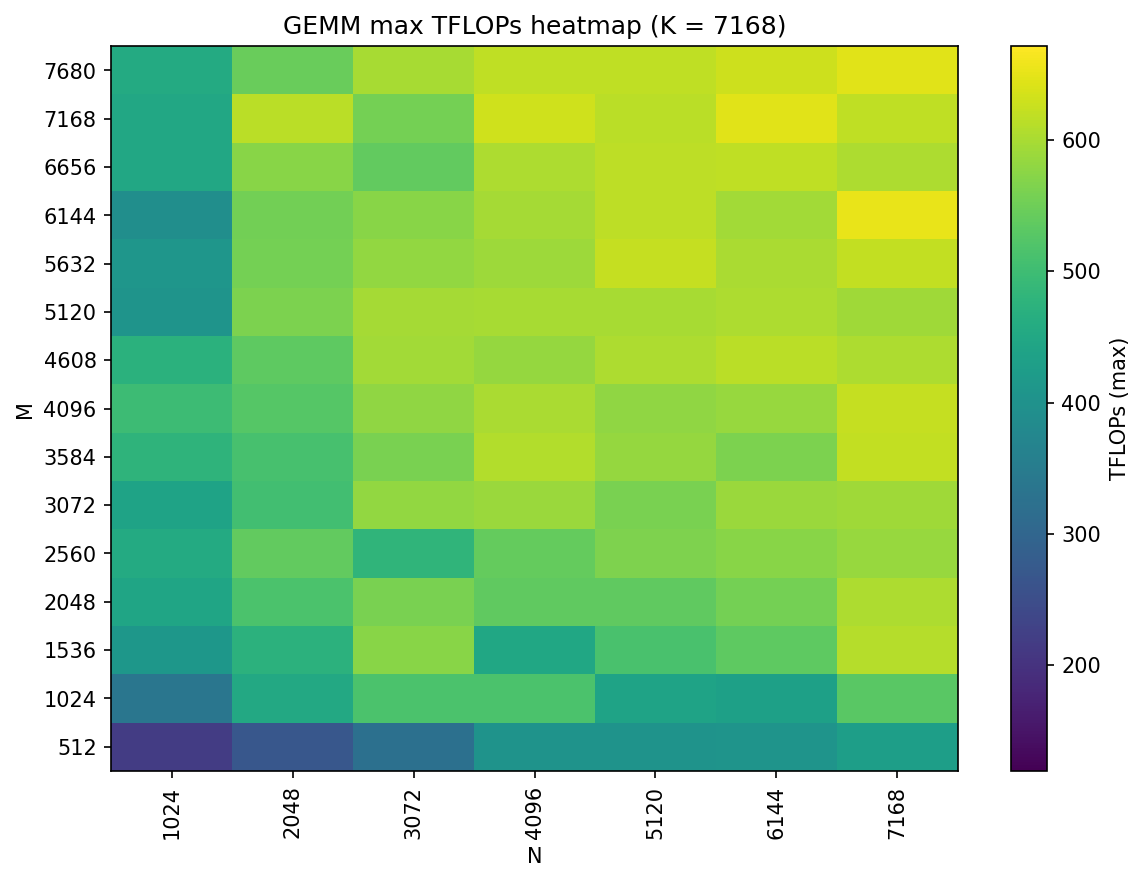}
        \caption*{(b) GEMM performance (TFLOPS/s) on MI300X with K=7168. Larger $K$ achieves higher throughput with smoother performance landscape. Peak performance reached at moderate problem sizes ($M, N \geq 512$).}
    \end{minipage}
    \caption{Heatmaps show achieved BFLOAT16 TFLOPS/s for (M, K) × (K, N) matrix multiplications with (a) K=1024 and (b) K=7168. Small K requires larger outer dimensions for efficiency, while large K achieves peak performance at moderate sizes.}
    \label{fig:sizing-heatmap}
\end{figure*}

Most of a model's iteration time is spent performing BF16 GEMM\footnotetext{GEMM (General Matrix Multiplication) computes $C = \alpha AB + \beta C$, where $A$ is an $M \times K$ matrix, $B$ is a $K \times N$ matrix, and $C$ is an $M \times N$ matrix, with $\alpha$ and $\beta$ as scalar coefficients. The operation requires $2MNK$ FLOPs (floating-point operations), as each of the $MN$ output elements involves a dot product of $K$ elements.} kernels (see Figure~\ref{fig:iter-breakdowns}). However, not all GEMM shapes are created equal—some shapes from a given model are far more amenable to a GPU's hardware than others (see Table~\ref{tab:ZAYA_gemms_smoe} and \citep{anthony2024codesign}). This makes sizing the model appropriately, so that the underlying GEMM shapes are performant on the target inference hardware, an extremely important step in model design.
To optimize GEMM performance for ZAYA1, we adopted a systematic approach. We began by performing an exhaustive search over all possible M, N, and K matrix shapes and their associated GEMM libraries (see Figure~\ref{fig:sizing-heatmap}). Beyond shape sizing, we also addressed the selection of GEMM backends and algorithms. ROCm provides multiple GEMM backends (rocBLAS and hipBLASLt), each containing many algorithms. We performed static tuning via a combination of PyTorch TunableOp, ROCm TransformerEngine \citep{transformer_engine}, and HIPBLASLt-bench tune from Primus \citep{primus2025}. This tuning produced static lookup tables that map GEMM sizes to the most performant algorithms within rocBLAS and hipBLASlt, which are then loaded at runtime to ensure optimal algorithm selection.
We then conducted a series of sizing sweeps across each component block in our model, allowing us to round the hyperparameters chosen for ZAYA1 (see Table~\ref{tab:ZAYA_varnames_smoe}) to efficient sizes. Our findings confirmed that larger GEMMs perform better—on the MI300X, a problem size of approximately 200 GFLOPs is required to reach peak throughput.

\subsection{InfinityFabric Bandwidth}
\label{sec:infinityfabric-bandwidth}

Since the growth in compute per GPU has far outpaced inter-node interconnect bandwidth per node, DL clusters require intra-node high-bandwidth interconnects such as AMD InfinityFabric \citep{amd2024cdna3} or NVIDIA NVLINK \citep{nvidia_nvlink}. These high-bandwidth intra-node interconnects are necessary to support parallelism schemes that require high communication volume such as expert- and tensor-parallelism \citep{gshard,megatron,anthony2024comms}. This allows the inter-node interconnect to be used for less bandwidth-intensive communication operations such as data parallelism or pipeline-parallelism \citep{rajbhandari2020zero, gpipe}. Additionally, high-bandwidth intra-node interconnects enable collective communication libraries (e.g. RCCL/NCCL/MPI) to design two-level collective algorithms that rely heavily on the intra-node fabric, thus indirectly enabling scale-out performance on the inter-node fabric.


Most DL software makes the assumption of a switched intra-node topology like NVSwitch. This topology allows for full bandwidth for any world size in a given communication operation. However, AMD's InfinityFabric uses xGMI \citep{xgmi}, which requires all GPUs participate in a given collective operation to achieve full bandwidth (see Eq. \ref{eq:intra-node-bw}). Parallelism schemes and DL implementations for current AMD clusters must therefore account for this. Our initial approach is that parallelism degrees should either span a full node or not be used at all. By enforcing this requirement, we can reformulate classically point-to-point schemes as collective operations. An example of this is tree attention~\citep{shyam2024tree}, which reformulates the attention operation such that the point-to-point message passing approach of ring-attention \citep{liu2023ring} can be replaced by AllReduce, an All-to-All operation.

We can formalize the achievable intra-node bandwidth using xGMI mathematically as follows,

\begin{equation}
\label{eq:intra-node-bw}
    B_{\text{per-GPU}} = \begin{cases}
    B_{\text{max}} & \text{NVSwitch (NVIDIA)} \\
    (n-1) \cdot B_{\text{link}} & \text{xGMI (AMD)}
\end{cases}
\end{equation}

Where $n$ is the number of GPUs participating in the intra-node communication operation ($1\leq n \leq 8$), $B_{\text{max}}$ is the maximum achievable bandwidth ($\approx 450$\,GBps for 8-MI300X communication across all xGMI links), and $B_{link}$ is the bandwidth of each xGMI link ($64$\,GBps for our MI300X xGMI fabric). 

\begin{figure*}[htbp]
    \centering
    \begin{minipage}[b]{0.32\textwidth}
        \centering
        \includegraphics[width=\textwidth]{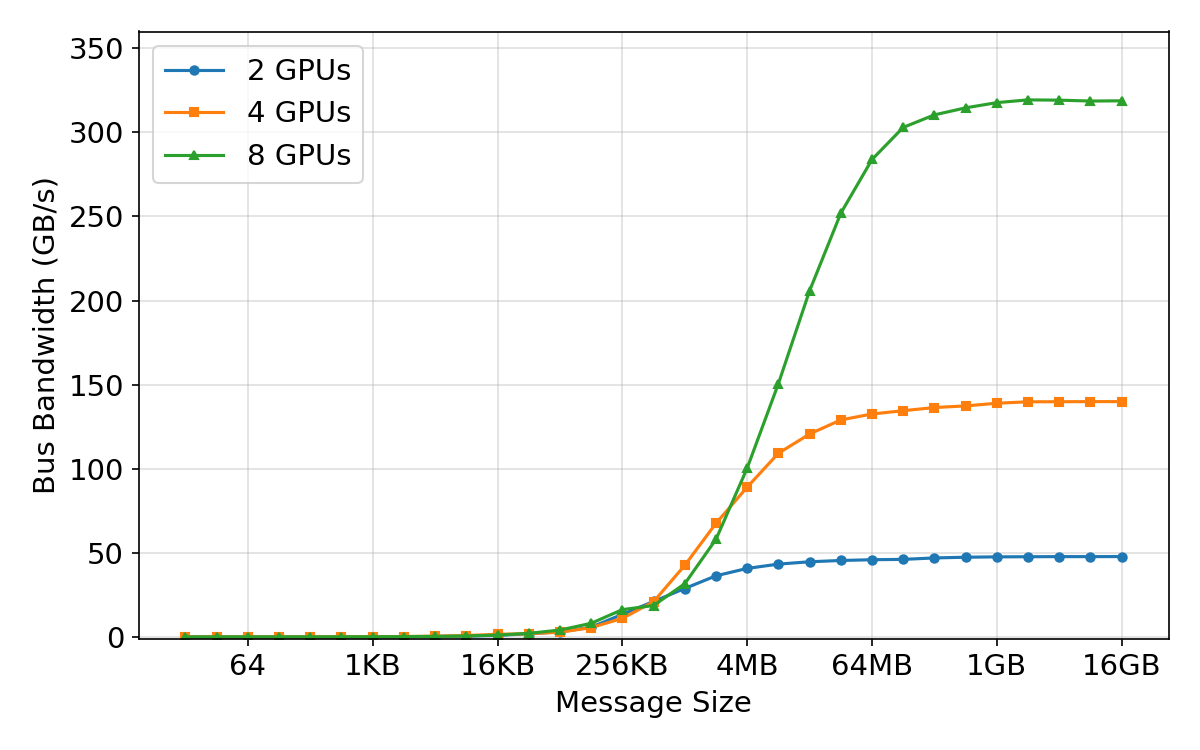}
        \caption*{(a) AllReduce Bus Bandwidth}
    \end{minipage}
    \hfill
    \begin{minipage}[b]{0.32\textwidth}
        \centering
        \includegraphics[width=\textwidth]{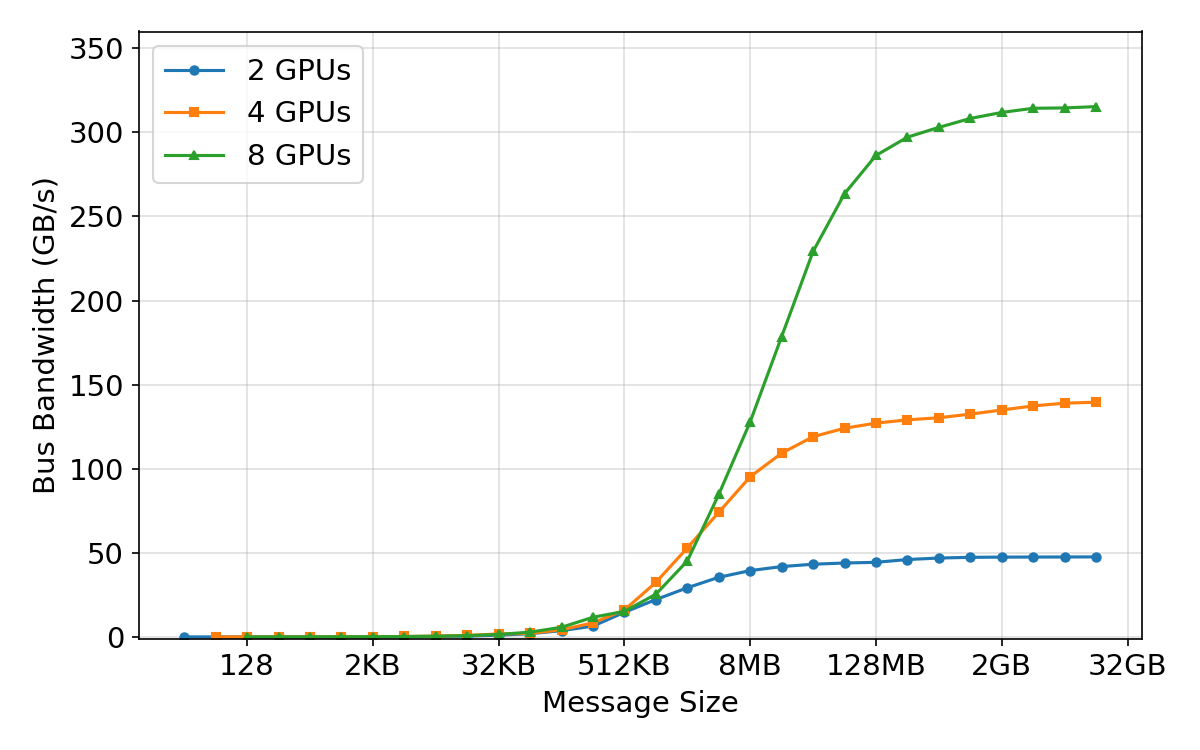}
        \caption*{(b) AllGather Bus Bandwidth}
    \end{minipage}
    \hfill
    \begin{minipage}[b]{0.32\textwidth}
        \centering
        \includegraphics[width=\textwidth]{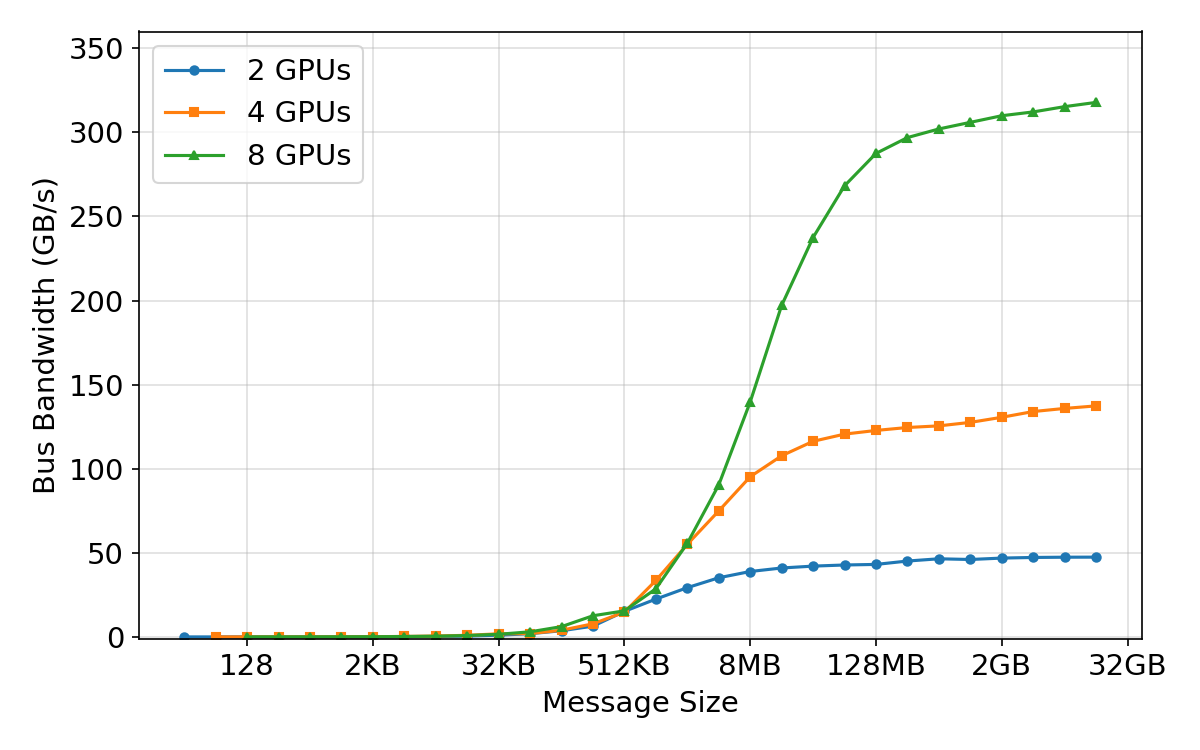}
        \caption*{(c) ReduceScatter Bus Bandwidth}
    \end{minipage}
    
    \vspace{0.5cm}
    
    \begin{minipage}[b]{0.32\textwidth}
        \centering
        \includegraphics[width=\textwidth]{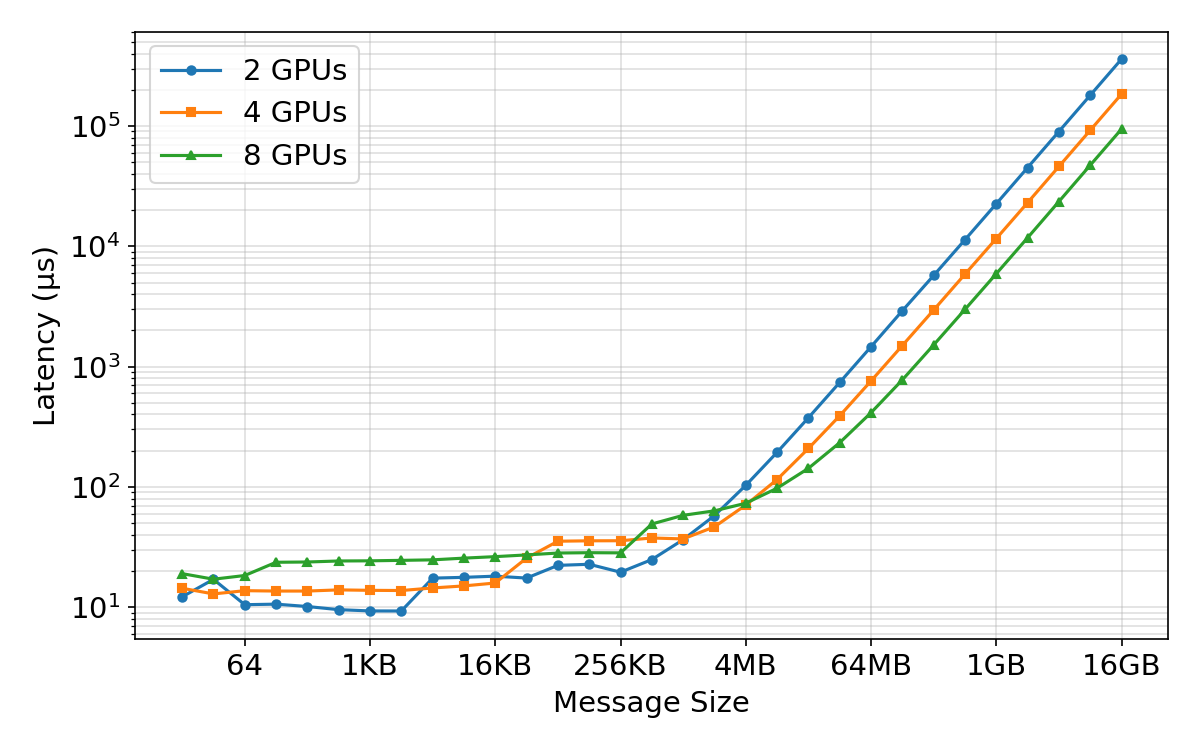}
        \caption*{(d) AllReduce Latency}
    \end{minipage}
    \hfill
    \begin{minipage}[b]{0.32\textwidth}
        \centering
        \includegraphics[width=\textwidth]{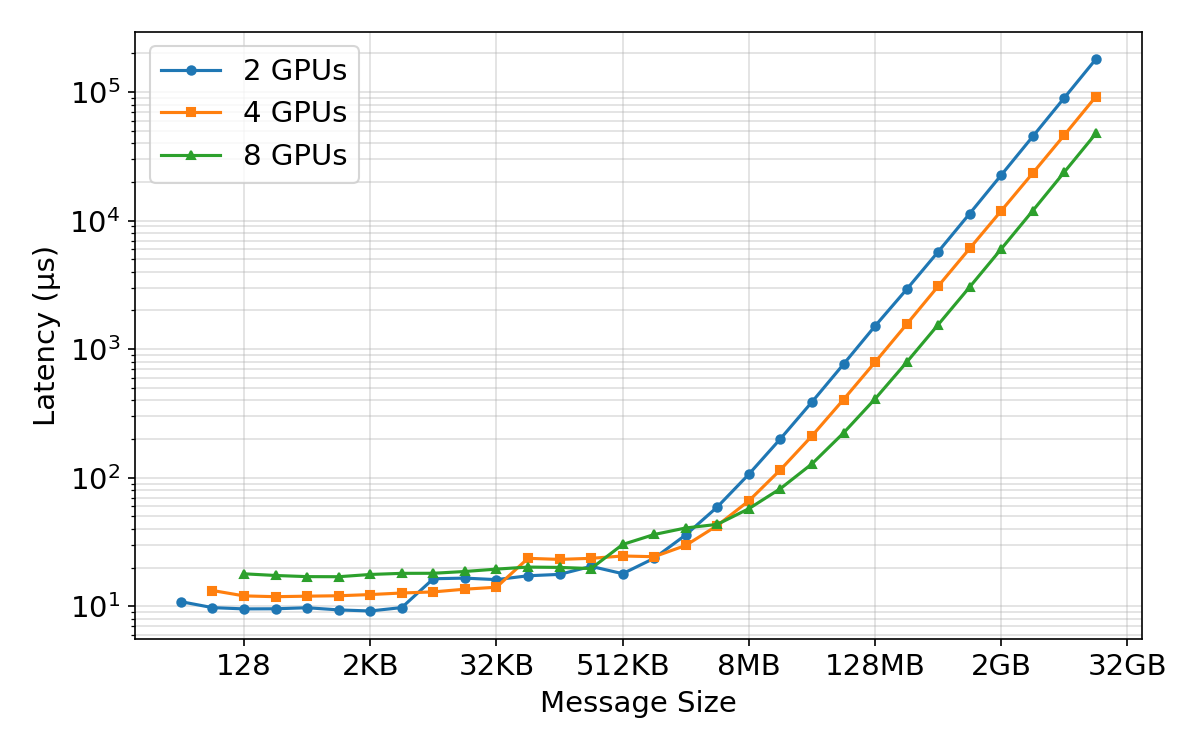}
        \caption*{(e) AllGather Latency}
    \end{minipage}
    \hfill
    \begin{minipage}[b]{0.32\textwidth}
        \centering
        \includegraphics[width=\textwidth]{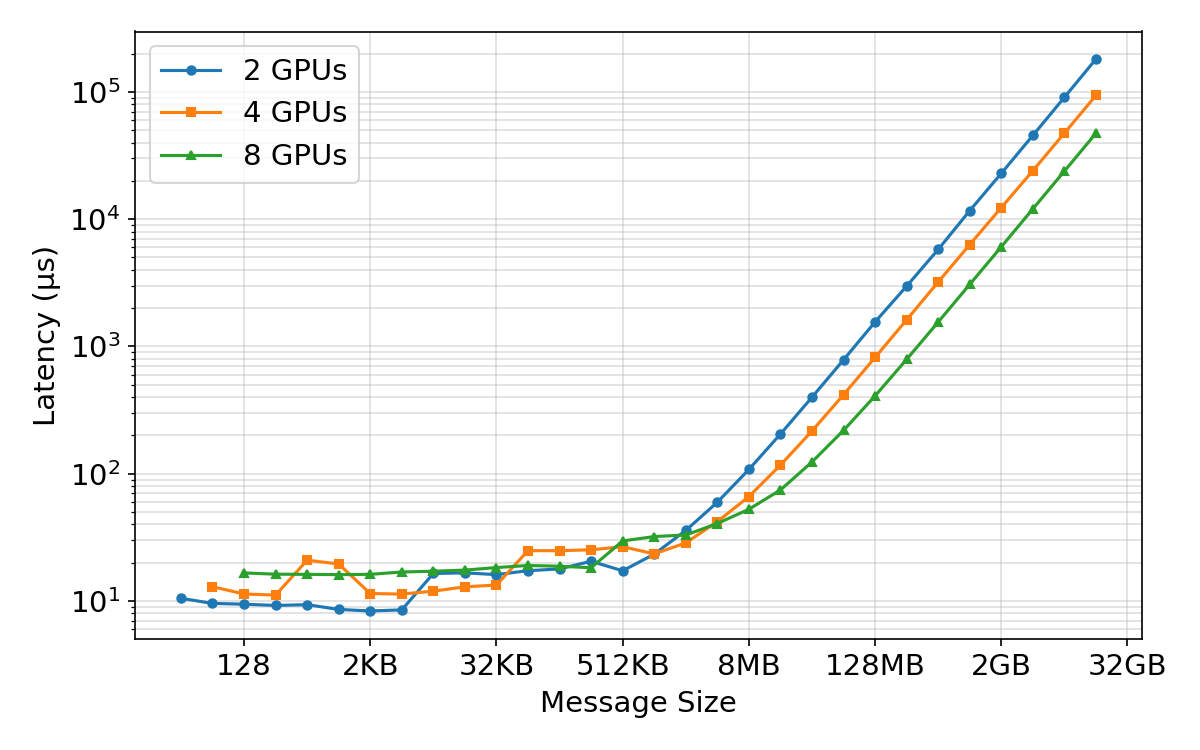}
        \caption*{(f) ReduceScatter Latency}
    \end{minipage}
    
    \caption{RCCL collective operations performance across InfinityFabric within a node. Top row shows bus bandwidth, bottom row shows latency for AllReduce, AllGather, and ReduceScatter operations.}
    \label{fig:rccl-collectives-intranode}
\end{figure*}

We depict results for intra-node communication operations in Figures \ref{fig:rccl-collectives-intranode}. Our parallelism topology for ZAYA1 is just the ZeRO-1 distributed optimizer during initial pretraining at 4096 sequence length, and ZeRO-1 + context-parallelism (ring attention \citep{liu2023ring}) when scaling context up to 32768. While the AllGather and AllReduce (ZeRO-1) operations were two-level algorithms, we incurred many point-to-point operations within the node when performing context-parallelism during pretraining. Long-context inference relies heavily on tree attention \citep{shyam2024tree} to avoid the bandwidth bottleneck of xGMI. This is because backwards compute will not be available for additional overlap, and forwards compute is not sufficient for full point-to-point overlap on xGMI.

\begin{figure*}[htbp]
    \centering
    \begin{minipage}[b]{0.42\textwidth}
        \centering
        \includegraphics[width=\textwidth]{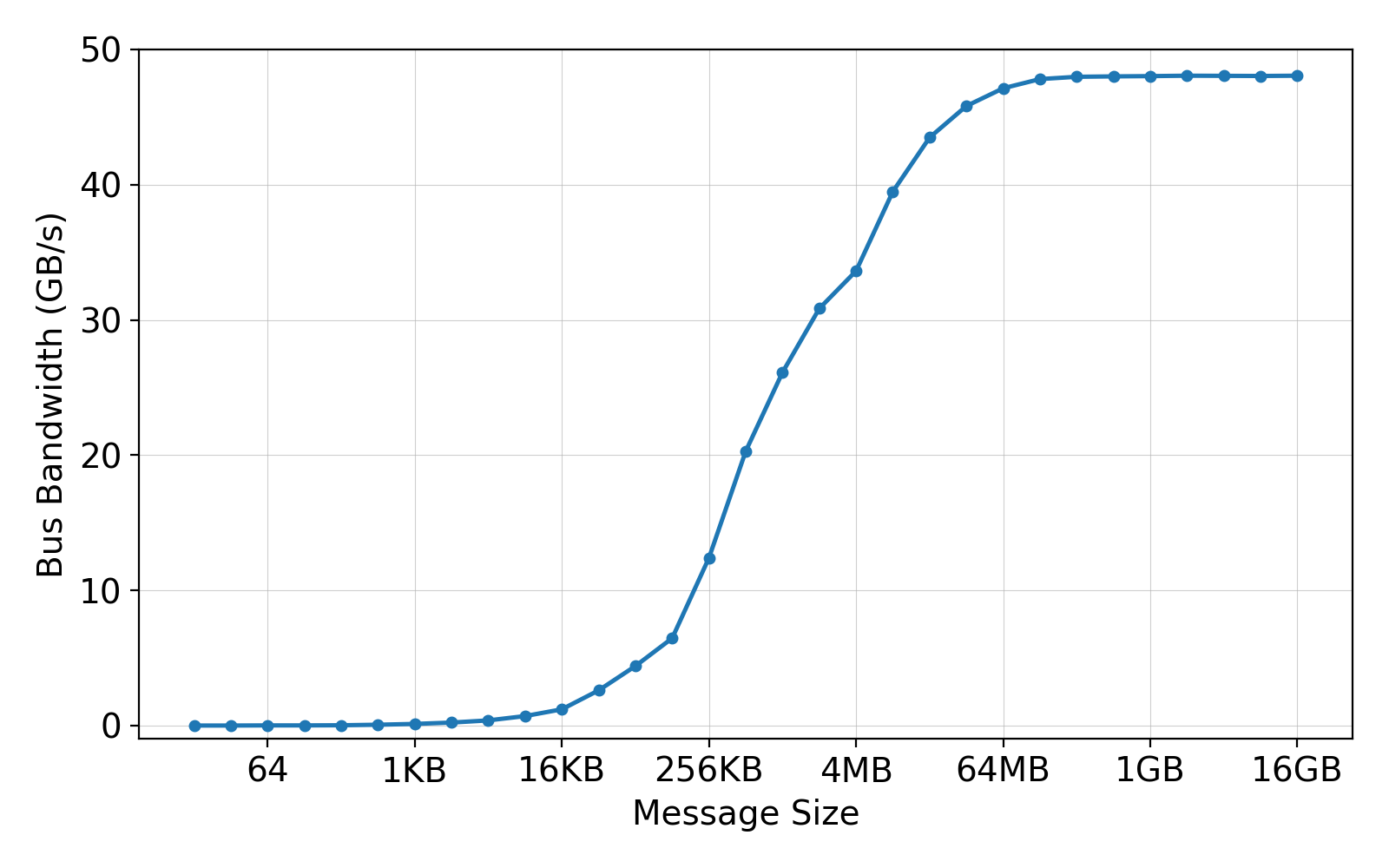}
        \caption*{(a) Bus bandwidth}
    \end{minipage}
    \begin{minipage}[b]{0.42\textwidth}
        \centering
        \includegraphics[width=\textwidth]{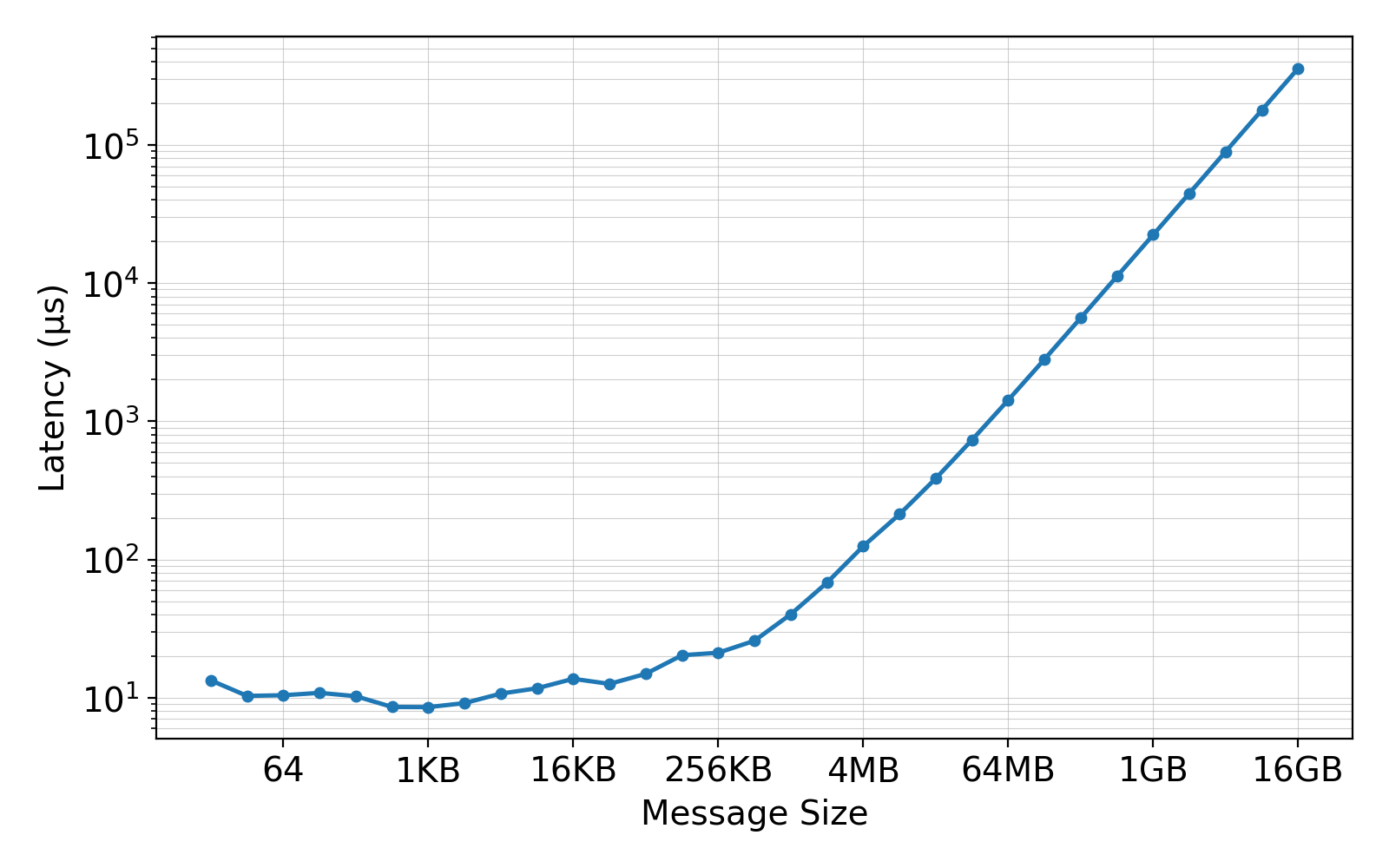}
        \caption*{(b) Latency}
    \end{minipage}
    \caption{RCCL Point-to-Point across InfinityFabric within a node}
    \label{fig:rccl-pt2pt-intranode}
\end{figure*}

\subsection{Pollara Bandwidth}
\label{sec:pollara-bandwidth}

Our cluster uses eight of the Pollara 400Gbps interconnects \citep{amd_pollara_2024} per compute node (see Figure~\ref{fig:cluster-arch}). While traditional HPC networking relied on low-latency across single- or dual-NIC topologies, communicating gradients during inter-node pretraining is inherently bottlenecked by the ratio of aggregate GPU throughput per node to the aggregate interconnect bandwidth per node. The full 3.2Tbps node bandwidth inherent in this cluster's nodes can be comfortably overlapped with compute unless models are extremely small. Further, each GPU having a dedicated NIC enables cost-efficient networking topologies such as our rails-only setup, reduces network congestion, and simplifies low-level communication software that heavily relies upon GPU-direct RDMA.

\begin{figure*}[htbp]
    \centering
    \begin{minipage}[b]{0.32\textwidth}
        \centering
        \includegraphics[width=\textwidth]{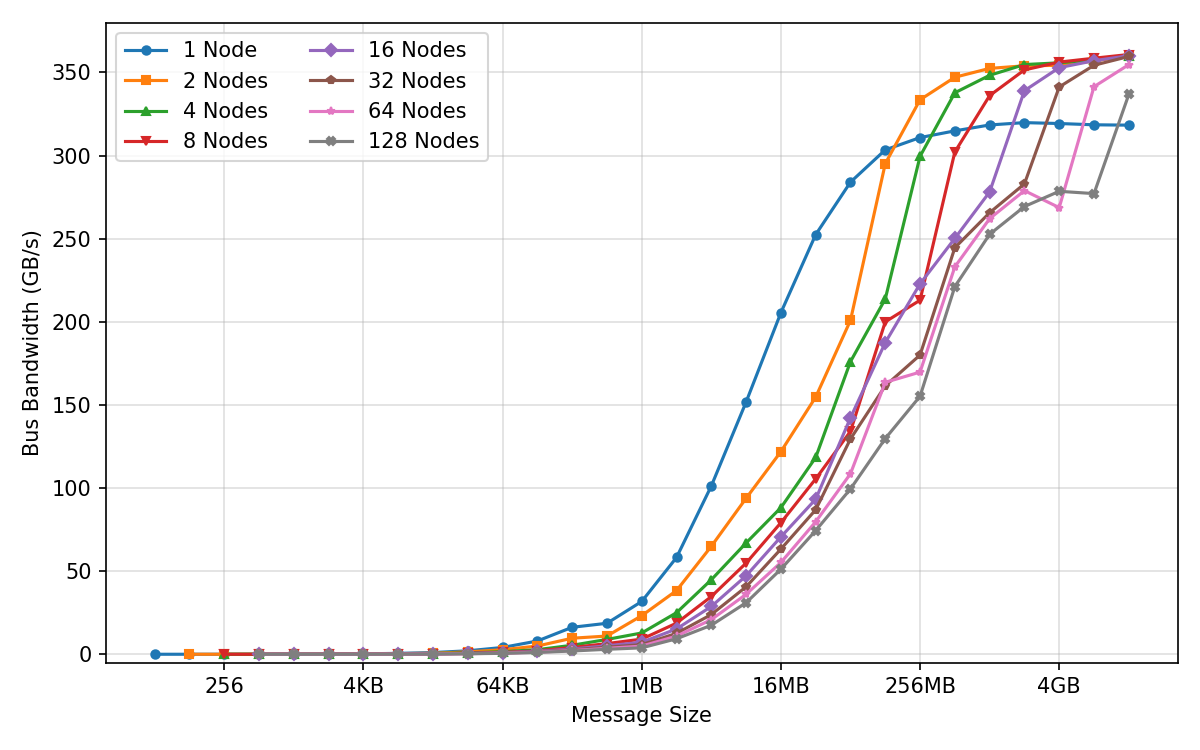}
        \caption*{(a) AllReduce Bus Bandwidth}
    \end{minipage}
    \hfill
    \begin{minipage}[b]{0.32\textwidth}
        \centering
        \includegraphics[width=\textwidth]{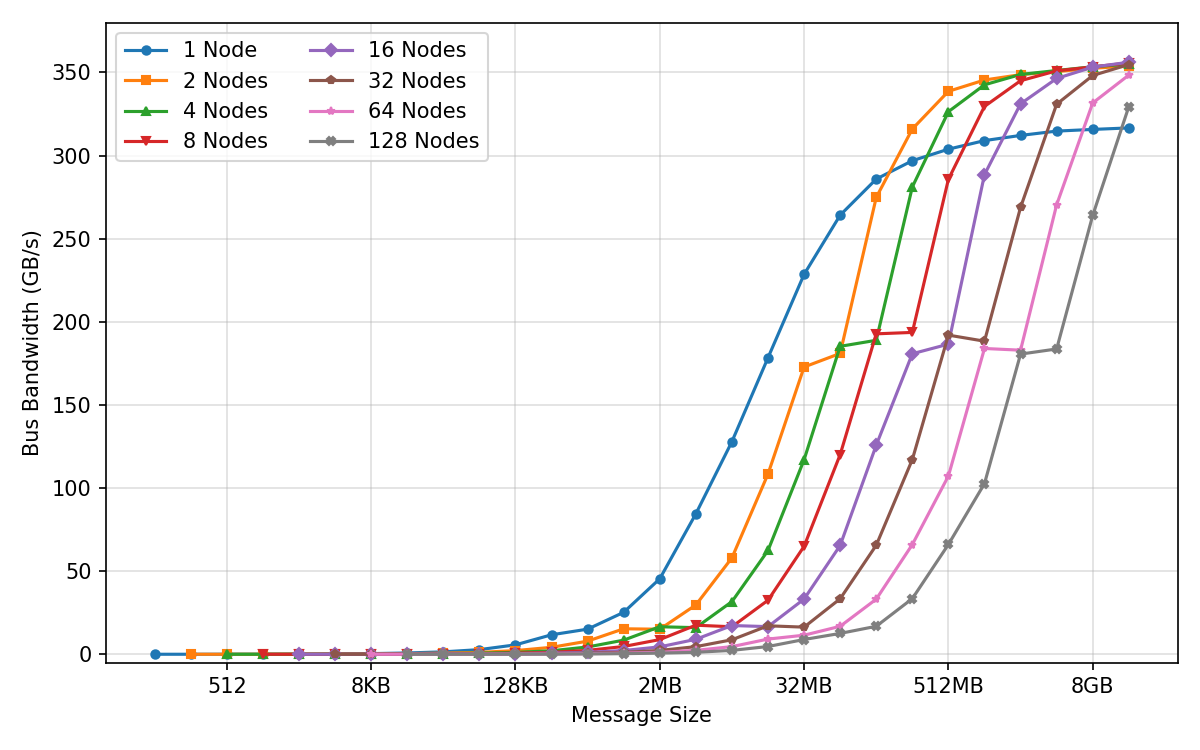}
        \caption*{(b) AllGather Bus Bandwidth}
    \end{minipage}
    \hfill
    \begin{minipage}[b]{0.32\textwidth}
        \centering
        \includegraphics[width=\textwidth]{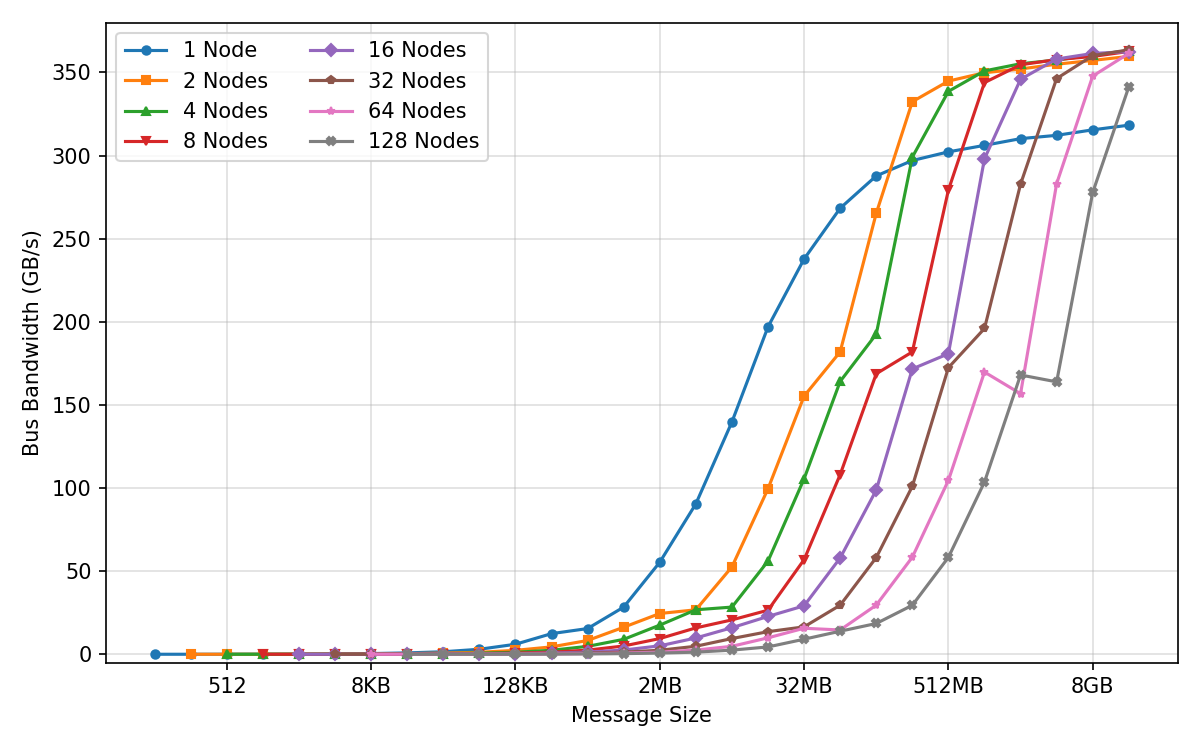}
        \caption*{(c) ReduceScatter Bus Bandwidth}
    \end{minipage}
    
    \vspace{0.5cm}
    
    \begin{minipage}[b]{0.32\textwidth}
        \centering
        \includegraphics[width=\textwidth]{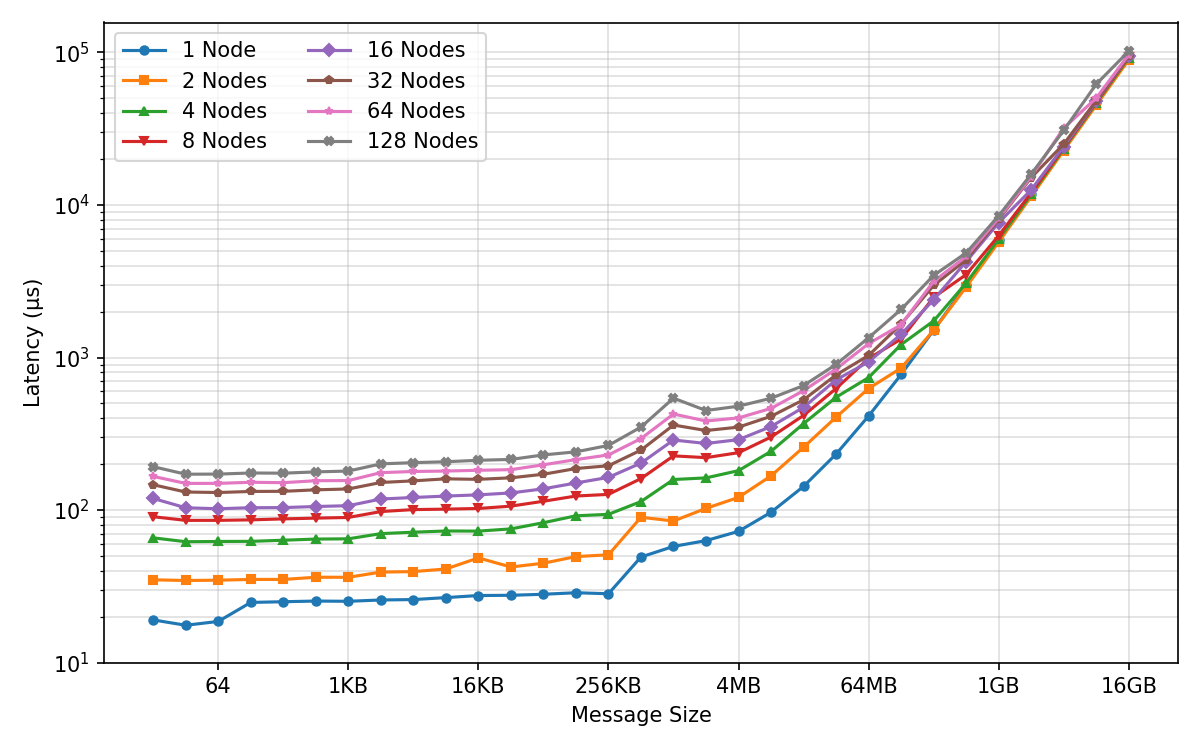}
        \caption*{(d) AllReduce Latency}
    \end{minipage}
    \hfill
    \begin{minipage}[b]{0.32\textwidth}
        \centering
        \includegraphics[width=\textwidth]{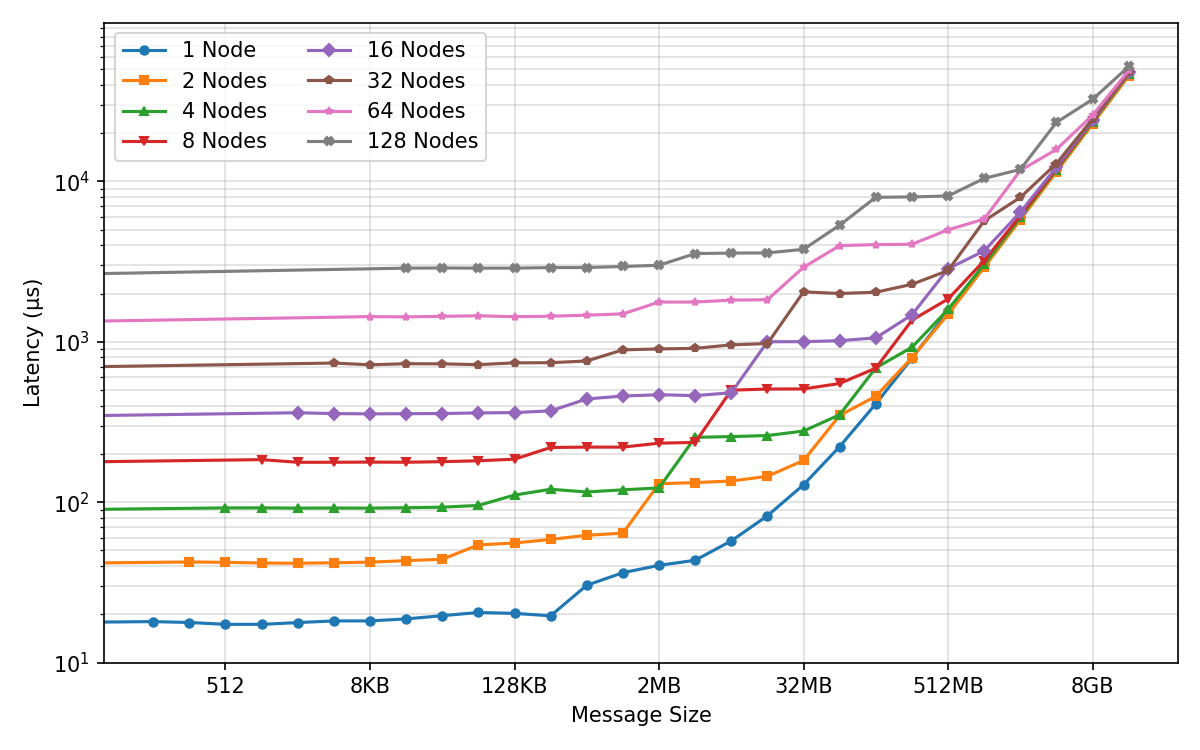}
        \caption*{(e) AllGather Latency}
    \end{minipage}
    \hfill
    \begin{minipage}[b]{0.32\textwidth}
        \centering
        \includegraphics[width=\textwidth]{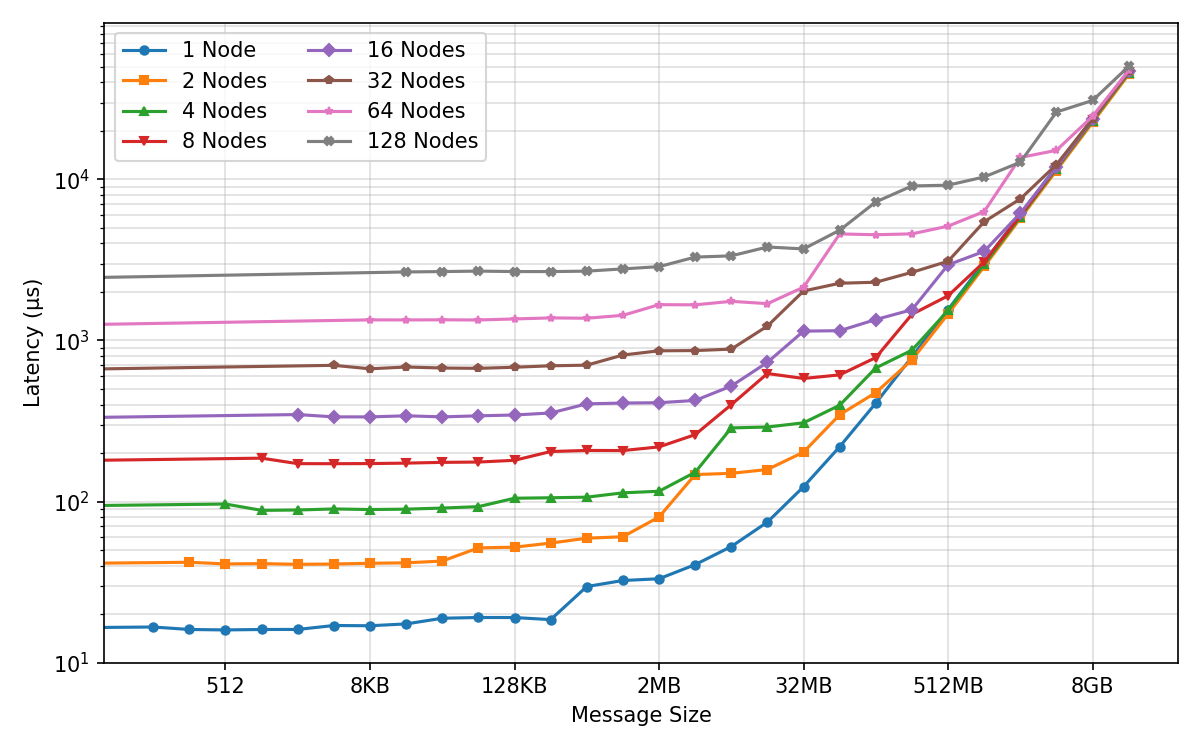}
        \caption*{(f) ReduceScatter Latency}
    \end{minipage}
    
    \caption{RCCL collective operations performance scaling across multiple nodes. Top row shows bus bandwidth, bottom row shows latency for AllReduce, AllGather, and ReduceScatter operations with varying node counts.}
    \label{fig:rccl-collectives-internode}
\end{figure*}

However, interconnects and their associated communication libraries require large messages to fully utilize the interconnect's bandwidth. Small messages are latency-bound (meaning that they spend the majority of their time in fixed per-message overheads such as kernel launch, synchronization, switch traversal, etc). Larger messages are bandwidth-bound (meaning that the one-time costs are amortized, and more bandwidth-efficient communication algorithms such as message pipelining may be used). See Figures~\ref{fig:rccl-pt2pt-intranode} and \ref{fig:rccl-collectives-internode} for examples of this effect.

During data-parallel training across nodes, we seek to communicate the full gradient tensor as quickly as possible and with as high a fraction of overlap with computation as possible. To this end, the message size of the communication operation should be large enough to reach the saturation point of the bandwidth figures in Figure~\ref{fig:rccl-collectives-internode}. However, making them any larger makes overlap more difficult since there are now larger and fewer communication operations to overlap with fixed-size computation operations. Distributed training frameworks such as ours provide a \textit{fusion buffer} to batch gradient tensors into, and then communicate the fused buffer. This provides practitioners with control over the size of the buffer to be communicated. We therefore set the fusion buffer size to rest exactly on this saturation point (but no larger) for ZAYA1 training.


\section{Model}
\label{sec:model}

\subsection{Architecture}

Our ZAYA1-base model utilizes a novel architecture which includes three key innovations upon contemporary MoE models: (1) CCA for the attention block, (2) ZAYA1 router, and (3) residual scaling. These architectural innovations significantly improve the per-parameter perplexity of ZAYA1 vs the ``classical'' MoE model architectures \citep{shazeer2016outrageously,fedus2022switch} with MLA or GQA attention and a linear router \citep{deepseekmoe}. CCA also improves training speed vs GQA and MLA and significantly reduces the FLOPs required for prefill, while maintaining equivalent KV-cache compression rates.

\begin{figure}
    \centering
    \includegraphics[width=0.6\linewidth]{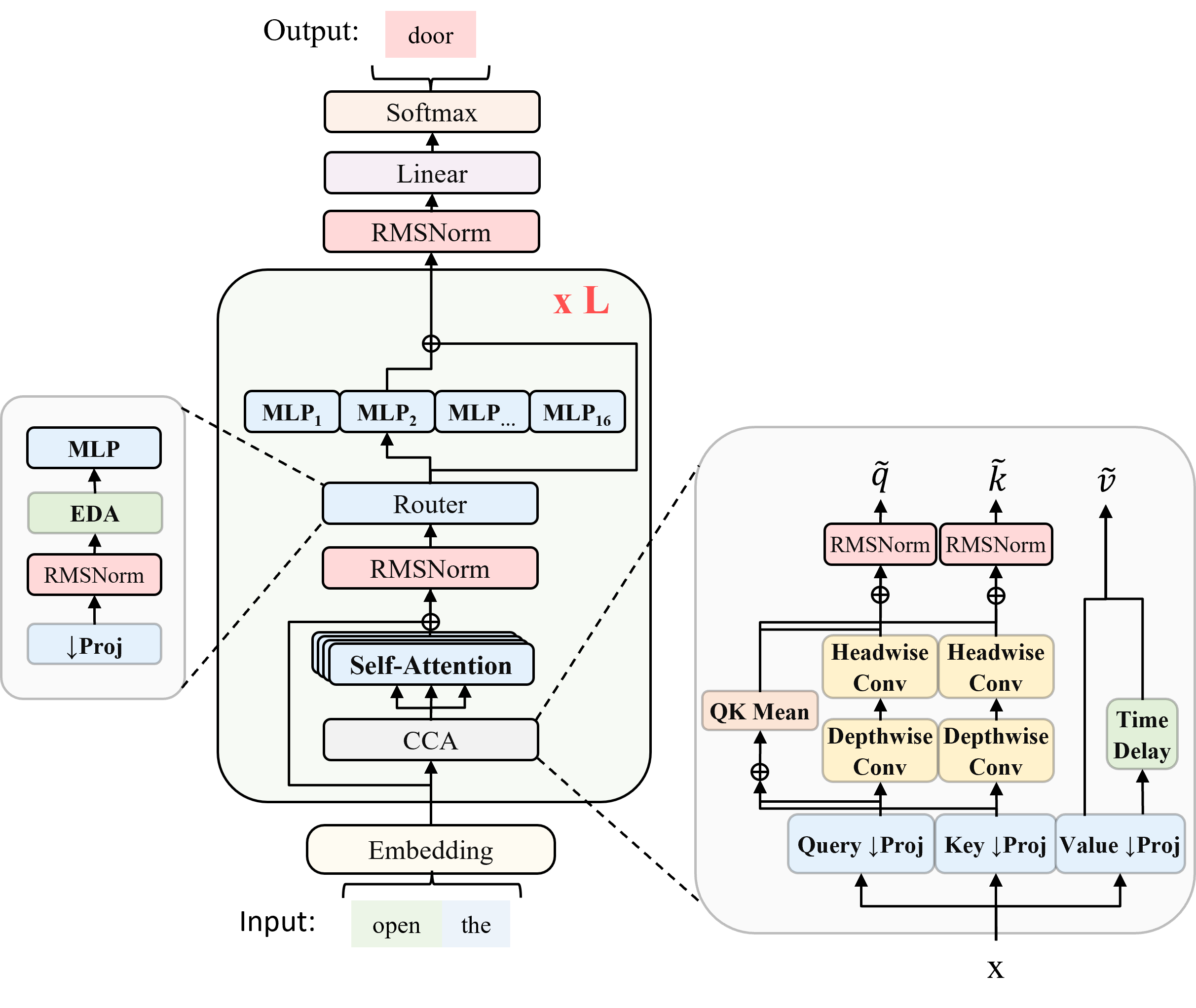}
    \caption{The model architecture of ZAYA1. The two core innovations in architecture presented here are CCA for the attention block and the ZAYA1 router. The ZAYA1 router replaces the linear router with a more expressive one consisting of downprojection, EDA, and then three sequential MLPs per expert.}
    \label{fig:ZAYA_arch}
\end{figure}

\subsubsection{Compressed Convolutional Attention (CCA)} CCA performs sequence-mixing entirely in a compressed latent space, allowing significant reductions in compute requirements for training and prefill, as well as large reductions in the KV cache. It matches other state of the art attention methods, such as MLA and GQA \citep{ainslie2023gqa,deepseekv3}. The performance of ZAYA1-base provides some evidence that CCA scales well and can support complex reasoning and in-context learning (ICL) behaviors as well as long-range recall. For more details on CCA see Appendix \ref{app:cca}. 

\subsubsection{ZAYA1 Router}
We make significant improvements to the standard linear router used in almost all large-scale MoE models. Firstly, we enhance the expressivity of the router by using an MLP in place of the standard linear router. Secondly, we mix the router's information with the routing choices of the previous layer's router using a mechanism we call \textit{Exponential Depth Averaging (EDA)}, since it represents an improvement to Depth-Weighted Averaging \citep{pagliardini2024denseformer}.

Given the residual stream input $x_l \in \mathbb{R}^{B \times S \times D}$, where $D$ is the residual stream dimension, the ZAYA1 router first down-projects the residual stream to a smaller router dimension $R$,
\begin{align}
    r_l = W_{\text{down}} x_l\,,
\end{align}
such that $r_l \in \mathbb{R}^{B \times S \times R}$. For ZAYA1-base we set $R=256$. 
We then apply EDA, which averages the representations with that of the previous layer, weighted by a learned coefficient $\gamma$:
\begin{align}
    r_l = r_l + \gamma r_{l-1}\,.
\end{align}
The EDA operation is followed by a three-layer MLP with GeLU activation functions to produce the final router scores $s \in \mathbb{R}^{B \times S \times E}$, where $E$ is the number of experts:
\begin{align}
    s_l = \text{softmax}(\text{MLP}(\text{RMSnorm}(r_l)))\,.
\end{align}
The scores are then used to select the chosen expert via top-k operation:
\begin{align}
    e_{\text{idx}} = \text{topk}(s_l + b_l)\,,
\end{align}
where $b_l$ are the learned vectors of bias balances. The ZAYA1 router uses an advanced bias balancing scheme which builds upon \citet{deepseekv3}. In our balancing method, the routing choices are learned using a scheme inspired by proportional–integral–derivative (PID) \citep{aastrom2006pid} controllers from classical control theory. This router enforces balancing across a microbatch. Our PID optimizer uses AdamW internally which we found to substantially improve the convergence speed and stability of the PID loop compared to the implementation presented in the literature.

We find that the increased expressivity of the MLP router and EDA can significantly improve the performance of MoE models as well as increasing ease of balancing and expert specialization. The additional MLPs in the router do require more FLOPs and a few additional parameters, however our parameter-matched ablations show that the router is an extremely strong place to add marginal parameters compared to the experts themselves or inside attention. Moreover, the number of parameters and FLOPs added inside the router is small due to the fact that the MLP operates in the downprojected latent space and not in the full embedding dimension. We find that the ZAYA1 router has a negligible impact on the iteration time of the model in practice. 

\subsubsection{ZAYA1 Residual Scaling}
The final architectural innovation in ZAYA1-base is residual scaling. We apply a learned bias $b_l$ and gating coefficient $\alpha \in \mathbb{R}^{D}$ both to the residual stream and to the input of each layer prior to RMSnorm:
\begin{align}
    x_{l+1} &= \text{Res-scale}_\alpha (x_l) + \text{Layer}(\text{RMSnorm}(\text{Res-scale}_\beta(x_l))) \\
    \text{Res-scale}_\alpha(x_l) &= \alpha x_l + b_l
\end{align}
Different gating coefficients and biases are applied to the residual stream and to the inputs to each layer. Residual scaling provides a method for the model to learn gating of unwanted inputs as well as to control the amount of ``forgetting'' that it can perform inside the residual stream. We observed that residual scaling achieves the same benefits as Qwen's proposed attention gating scheme \citep{qiu2025gated}, without any of the parameter and FLOP overhead required by an explicit gating matrix. Residual scaling also helps to control the growth of the residual norm through the network depth. 
Because residual scaling only adds $2 \times L \times D$ parameters to the network, its parameter and FLOP overhead are similar to layernorm and are negligible. 

Beyond these architectural innovations, we trained with 16 experts with a hidden dimension expansion factor of 2, thus moving towards fine-grainedness which, like others in the literature \citep{team2025kimi,deepseekai2024deepseekv32,deepseekmoe,tian2025towards}, we found improved performance at a fixed parameter sizing. 

Unlike many contemporary MoEs, we train with a topk of 1 and without residual experts \citep{rajbhandari2022deepspeed,deepseekv3}. We find that the improved routing expressiveness of the ZAYA1 router and resulting increasing specialization of the experts makes using a residual expert unnecessary. Moreover, we find that using higher top-ks is less effective than top-1 in FLOP-matched experiments when using the ZAYA1 router. We hypothesize that this is because the ZAYA1 router is more certain about the experts it picks, thus the additional experts added by top-k are less necessary and also that when larger ks are used, the contribution of these experts is diminished due to the multiplication by the routing probability. In general, we observe that ZAYA1 produces significantly lower entropy routing probabilities than linear routers, indicating high certainty in its choice. For attention, we utilized CCGQA with a query compression rate of $2\times$ and a KV compression rate of $8\times$. We applied RoPE \citep{su2023rotary} to half the channels in each head, leaving the other half without position embeddings. ZAYA1-base was trained using the Gemma3 tokenizer.

\subsection{Training}

\begin{figure}
    \centering
    \includegraphics[width=0.85\linewidth]{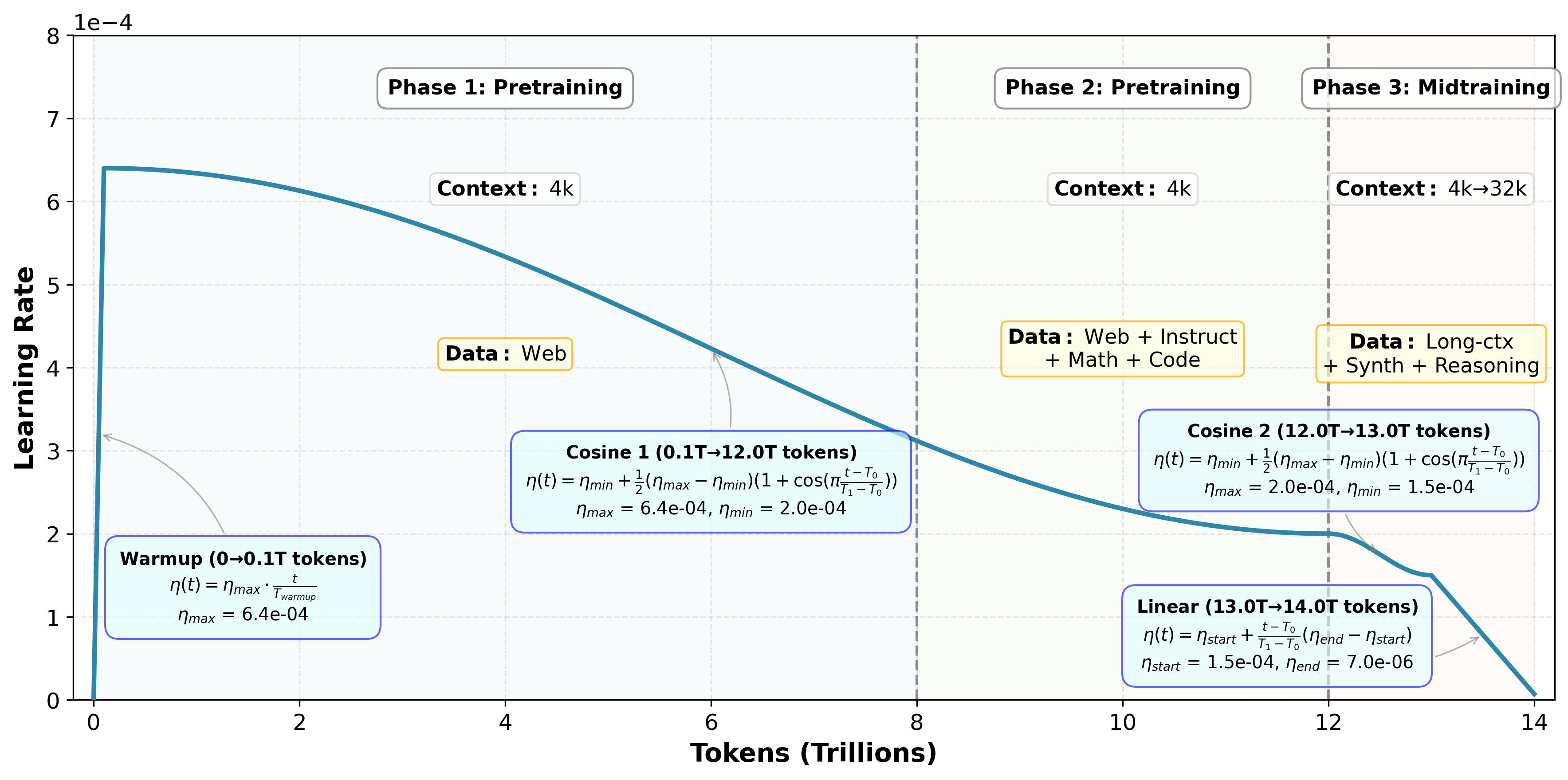}
    \caption{Schematic of the three phases of pretraining for ZAYA1-base. Data mixture, learning rate schedule, and context length are chosen for each phase so that the model is prepared for post-training. The core pretraining consists of two phases. The first phase inculcates general knowledge and linguistic understanding into the model through highly diverse corpora of primarily web-sourced data. The second phase begins to reinforce and strengthen the mathematics, coding, and STEM knowledge components through additional mixing of information-rich high-quality data. The final phase extends the context and further emphasizes STEM content as well as prepares the base for instruction-following and reasoning post-training.}
    \label{fig:schedule}
\end{figure}

Following recent works \citep{gu2025jet, bakouch2025smollm3}, ZAYA1-base was trained in three phases. During phase 1, which lasted for 8T tokens, ZAYA1-base was pretrained on a mix that consisted primarily of web-crawl data with additional code, math, and multilingual data mixed in. This was followed by a 4T-token phase 2 in which the proportion of code and math data was significantly increased. Further, additional reasoning and SFT-formatted data was added, with reasoning traces trimmed to preserve full answers within the 4k context window. During these two phases, the learning rate schedule was a cosine decay from 6e-4 to 2e-4. During the mid-training phase 3, we used a cosine decay from 2e-4 to 1.5e-4 during context extension. This was followed by our final annealing phase, for which we used a linear decay to 7e-6 to prepare for post-training. See Figure~\ref{fig:schedule} for our full training schedule to produce ZAYA1-base.

The mid-training phase 3 lasted for 1T tokens, during which we further increased the proportion of math, code, and reasoning data.  with special focus on reasoning data which could now be trained at full length without truncation. We also upweighted natively long context datasets such as books which, similar to others in the literature \citep{xu2025128k,mo2025mid,gao2024train,Yarn}, we found was important in achieving good long context performance. For context extension, we extended the RoPE base frequency from 10K to 1M while we progressively expanded the context from 4k to 32k. At the end of the context extension, we performed a mid-train at 32k context with a heavy focus on synthetic mathematical, code, and instruction following data. We additionally performed a 1.2T reasoning focused mid-train on synthetic long-CoT data from the checkpoint before annealing with a small amount of replay of the data distribution from phase 2 to prime the model for further post-training and RL.

Due to the large HBM capacity of the MI300x GPU, which stands at 192\,GB, we were able to pretrain ZAYA1-base without requiring complex parallelism strategies. We pretrained solely using data-parallelism with the ZeRO-1 distributed optimizer, where only the optimizer states are sharded. Adopting Muon, which only requires the momentum and not the variance term of AdamW, also reduced the memory requirements of training compared to AdamW.

During phase 1 of pretraining, we were able to fit a minibatch size of 5 in HBM. During context extension to 32k, we trained with minibatch size 1 and with context-parallel size of 2. For 128k context, we merely require training with a context-parallel size of 8 and do not yet require inter-node communication.
The substantial reductions in attention FLOPs and activation memory that CCA provided also made context extension significantly easier than for alternative attention methods since both our training FLOPs and activation memory were reduced by approximately $8\times$ compared to e.g. MLA. This allowed us to train at 32k context with approximately the same efficiency at 4k context.   

ZAYA1-base is trained using the Muon optimizer \citep{jordan6muon}. We performed five Newton-Schulz iterations per gradient step. We applied Muon to all 2D parameters except the embedding tables; all other parameters were trained with AdamW. We optimized the Q,K,V matrices of attention separately to increase the `squareness' of the weight matrices which improves Muon performance. We apply $0.2 \sqrt{\max(a,b)}$ to the Newton-Schulz update to rescale the update RMS-norm of Muon to the mean field estimate of the update RMS-norm of AdamW, as suggested by~\citet{liu2025muon}, allowing us to use a unified learning rate for the entire architecture. 

During training, we performed batch size scheduling, where the global batch size was increased from 16M tokens during phase 1 and phase 2 pretraining to 30M during the mid-training phase. Overall we found that both Muon and MoE models require larger batches for peak efficiency during pretraining \citep{shen2025convergence}. For MoE models this makes sense since each expert will only see (approximately) $\frac{1}{E}$ tokens where $E$ is the number of experts. For instance, ZAYA1 has 16 experts and so at a batch size of 16M each expert only sees 1M tokens, which is plausibly far below the critical batch for those expert parameters. The attention blocks are likely above their critical batch size, but the parameter efficiency of CCA makes this less harmful for training efficiency than it would otherwise be. Moreover, Muon in general appears to have higher critical batch size than AdamW. This tolerance for high batch sizes is an underrated aspect of MoE and Muon efficiency since increasing the global batch size is inevitable as compute cluster size grows, which limits the maximum size of the cluster on which it is efficient to train.

Our training framework is a heavily modified version of Megatron-LM \citep{megatron}; we integrated and utilized several components from AMD's Primus framework \citep{primus2025} such as monitoring tooling and some kernels to help ease the transition to AMD pretraining.

\section{Performance Optimization}
\label{sec:optimization}

\subsection{Model Sizing}
\label{sec:model-sizing}

\begin{table}[ht]
\centering
\resizebox{.7\columnwidth}{!}{%
\begin{tabular}{l|l|l}
\toprule
\textbf{Symbol} & \textbf{Definition} & \textbf{ZAYA1-base value} \\
\midrule
$\textit{a}$      & Number of attention heads                  & $16$ \\
$\textit{g}$      & Number of key/value heads ($g \le a$)      & $2$ \\
$a_q$             & Number of query heads used by attention (CCA) & $8$ \\
$c_q$             & Query head fraction $= a_q/a$              & $1/2$ \\
$c_{kv}$          & KV head fraction $= g/a$                   & $1/8$ \\
$\textit{b}$      & Microbatch size                            & $b$ ($5\rightarrow1$ as $s$ grows) \\
$\textit{s}$      & Sequence length                            & $s$ ($4096-32768$ for base model) \\
$\textit{t}$      & Tensor-parallel size                       & $t=1$ (no TP used) \\
$\textit{h}$      & Hidden dimension size                      & $2048$ \\
$d_h$             & Head dimension $= h/a$                     & $128$ \\
$\textit{L}$      & Number of transformer layers               & $40$ \\
$\textit{v}$      & Vocabulary size                            & $262272$ \\
$E_\ell$          & Experts in layer $\ell$ (local to rank)    & $16$ \\
$k$               & Router top-$k$ (experts per token)         & $1$ \\
$\tau_i$          & Tokens routed to local expert $i$          & $\tau_i\approx(\frac{s}{E_{\ell}}=256)$ \\
$D$               & Router expansion dimension                 & $256$ \\
$f$               & Expert FFN width (pre-activation)          & $4096$ \\
$f_o$             & Expert post-activation width               & $2048$ (SwiGLU; $f/2$) \\
$k_0,k_1$         & CCA conv kernel sizes along sequence        & $k_0=2, k_1=2$ \\
\bottomrule
\end{tabular}%
}
\caption{Symbols and instantiated values for the provided ZAYA1 configuration.}
\label{tab:ZAYA_varnames_smoe}
\end{table}

As found in \citep{anthony2024codesign}, we can achieve efficiency benefits if models are carefully sized so that the underlying kernels have amenable sizes to work with the specific GPU hardware chosen for training and inference.  This sizing sweep must be performed for every hardware that the model creators want to target, focusing primarily upon the pretraining and inference GPUs and associated parallelism topologies, since these steps constitute the most FLOP-cost \citep{mosaicml-chinchilla} within the lifetime of a given model. Typical targets for sizing include the hidden dimension, the head dimension, and the embedding sizes.

To this end, we ran a suite of sizing experiments targeting the GEMM (General Matrix Multiplication) kernels which comprise the model architecture that we seek to target. These operations and their associated sizes are provided below in Table~\ref{tab:ZAYA_gemms_smoe}. The model architecture that comprises ZAYA1-base therefore has the following sizing constraints:

\begin{itemize}
    \item The vocabulary size $v$ should be divisible by $64$.
    \item The microbatch size $b$ should be as large as possible~\citep{nado2021large}.
    \item $b \cdot s$, $\frac{h}{a}$, and $\frac{h}{t}$ should be divisible by a power of two, though there is no further benefit to going beyond $64$.
    \item $(b \cdot a)/t$ should be an integer, for any of CCA's $a$.
    \item The degree of tensor- ($t$) and expert-parallelism should be as small as possible~\citep{narayanan2021efficient} \citep{comms-modeling}. Parallelism should only be used when HBM quantity is insufficient, and scales in latency with increasing world size and message size.
\end{itemize}

We have performed sizing sweeps for both raw GEMMs and the most sensitive operations such as flash attention. The results of some raw GEMM sweeps are in Section \ref{sec:gpu-compute}. One challenge to note here is that the GEMM input sizes for MoE models are dynamic, and depend on the state of balancing. In order to avoid massive tuning tables that would only be picked up for extremely unbalanced cases, we define bands of expected sequence lengths around the expected perfectly-balanced input lengths per MLP of $[4096, 8192, 16384, 32768] / 16 = [256, 512, 1024, 2048]$ tokens as we extend context lengths. The band of tolerance around each sequence length is $50\%$, meaning that we tune within 128-384 tokens for the 4096 sequence length pretraining phase.

\begin{table}[ht]
\centering
\resizebox{\columnwidth}{!}{%
\begin{tabular}{lccc}
\toprule
\textbf{Module / Operation} & \textbf{GEMM Size (per rank)} & \textbf{Notes / Kernel mapping} & \textbf{ZAYA1-base (per rank)} \\
\midrule
Input Embedding                       & --- & lookup / gather & --- \\
Fused Add+Norm (pre-attn)             & --- & fused add + (RMS)Norm kernel & --- \\
\midrule
CCA $Q$ projection (CCA layers)       & $(b\!\cdot\! s,\ h)\ \times\ \big(h,\ \tfrac{a_q\,(h/a)}{t}\big)$ & GEMM (column-parallel)
                                      & $(b\!\cdot\! s,\,2048)\ \times\ (2048,\,\tfrac{1024}{t})$ \\
CCA $K$ projection (CCA layers)       & $(b\!\cdot\! s,\ h)\ \times\ \big(h,\ \tfrac{g\,(h/a)}{t}\big)$   & GEMM (column-parallel)
                                      & $(b\!\cdot\! s,\,2048)\ \times\ (2048,\,\tfrac{256}{t})$ \\
CCA $V_1$ projection (CCA layers)     & $(b\!\cdot\! s,\ h)\ \times\ \big(h,\ \tfrac{a_k d}{2t}\big)$     & GEMM (first value stream)
                                      & $(b\!\cdot\! s,\,2048)\ \times\ (2048,\,\tfrac{128}{t})$ \\
CCA $V_2$ projection (CCA layers)     & $(b\!\cdot\! s,\ h)\ \times\ \big(h,\ \tfrac{a_k d}{2t}\big)$     & GEMM (delayed value stream)
                                      & $(b\!\cdot\! s,\,2048)\ \times\ (2048,\,\tfrac{128}{t})$ \\
CCA depthwise Conv1d                  & --- & depthwise Conv1d (kernel $k_0$) & --- \\
CCA grouped Conv1d                    & --- & grouped Conv1d (groups $a_q{+}g$, kernel $k_1$) & --- \\
RoPE                                   & --- & elementwise rotate & --- \\
Repeat-KV (GQA expand)                 & --- & repeat/interleave (memcpy/index) & --- \\
\midrule
Attention score $QK^\top$ (CCA)       & $\big(\tfrac{b\,a_q}{t},\ s,\ \tfrac{h}{a}\big)\ \times\ \big(\tfrac{b\,a_q}{t},\ \tfrac{h}{a},\ s\big)$ & Flash Attention \citep{dao2023flashattention2}
                                      & $\big(\tfrac{8b}{t},\, s,\,128\big)\ \times\ \big(\tfrac{8b}{t},\,128,\, s\big)$ \\
Attention apply to $V$ (CCA)          & $\big(\tfrac{b\,a_q}{t},\ s,\ s\big)\ \times\ \big(\tfrac{b\,a_q}{t},\ s,\ \tfrac{h}{a}\big)$           & Flash Attention \citep{dao2023flashattention2}
                                      & $\big(\tfrac{8b}{t},\, s,\, s\big)\ \times\ \big(\tfrac{8b}{t},\, s,\,128\big)$ \\
Attn output projection $O$ (CCA)      & $\big(b\!\cdot\! s,\ \tfrac{a_q}{a}\tfrac{h}{t}\big)\ \times\ \big(\tfrac{a_q}{a}\tfrac{h}{t},\ h\big)$ & GEMM (row-parallel)
                                      & $(b\!\cdot\! s,\,\tfrac{1024}{t})\ \times\ (\tfrac{1024}{t},\,2048)$ \\
Fused Add+Norm (post-attn)            & --- & fused add + (RMS)Norm kernel & --- \\
\midrule
Router down-proj (MoE layers)         & $(b\!\cdot\! s,\ h)\ \times\ (h,\ D)$                               & GEMM
                                      & $(b\!\cdot\! s,\,2048)\ \times\ (2048,\,256)$ \\
Router MLP 1                          & $(b\!\cdot\! s,\ D)\ \times\ (D,\ D)$                                & GEMM + GELU
                                      & $(b\!\cdot\! s,\,256)\ \times\ (256,\,256)$ \\
Router MLP 2                          & $(b\!\cdot\! s,\ D)\ \times\ (D,\ D)$                                & GEMM + GELU
                                      & $(b\!\cdot\! s,\,256)\ \times\ (256,\,256)$ \\
Router logits                         & $(b\!\cdot\! s,\ D)\ \times\ (D,\ E_\ell)$                           & GEMM; $E_\ell$ local experts (+1 if using MOD)
                                      & $(b\!\cdot\! s,\,256)\ \times\ (256,\,16)$ \\
Top-$k$ select \& probs               & --- & softmax + \texttt{topk} & --- \\
Token dispatch / (permute, scatter)   & --- & sort/gather/scatter kernels & --- \\
\midrule
Expert MLP fc1 (per expert $i$)       & $(\tau_i,\ h)\ \times\ (h,\ f)$                                      & GEMM; grouped/batched across local experts
                                      & $(\tau_i,\,2048)\ \times\ (2048,\,4096)$ \\
Expert activation (GeGLU/SwiGLU)      & --- & fused bias+act elementwise & --- \\
Expert MLP fc2 (per expert $i$)       & $(\tau_i,\ f_o)\ \times\ (f_o,\ h)$                                  & GEMM; $f_o$ as in Table~\ref{tab:ZAYA_varnames_smoe}
                                      & $(\tau_i,\,2048)\ \times\ (2048,\,2048)$ \\
Combine \& inverse permute            & --- & gather/scatter back & --- \\
Fused Add+Norm (post-MLP)             & --- & fused add + (RMS)Norm kernel & --- \\
\midrule
Final Norm                            & --- & (RMS)Norm kernel & --- \\
Linear Output (LM head)               & $(b\!\cdot\! s,\ h)\ \times\ (h,\ v)$                                & GEMM (ties to embedding if shared)
                                      & $(b\!\cdot\! s,\,2048)\ \times\ (2048,\,262272)$ \\
\bottomrule
\end{tabular}%
}
\caption{ZAYA1-base operators and GEMM sizes}
\label{tab:ZAYA_gemms_smoe}
\end{table}

\begin{figure*}[htbp]
    \centering
    \begin{minipage}[b]{0.42\textwidth}
        \centering
        \includegraphics[width=\textwidth]{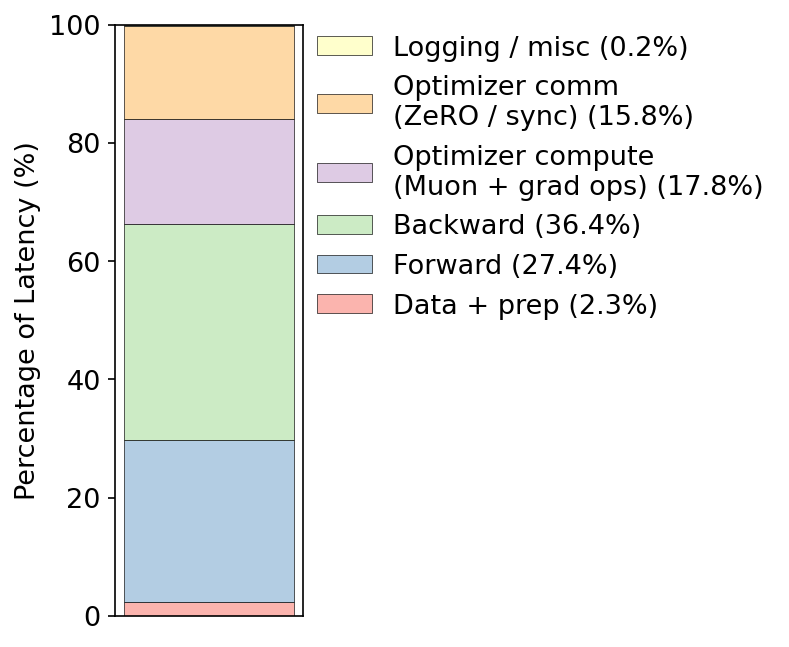}
        \caption*{(a) Training iteration latency breakdown}
    \end{minipage}
    \begin{minipage}[b]{0.42\textwidth}
        \centering
        \includegraphics[width=\textwidth]{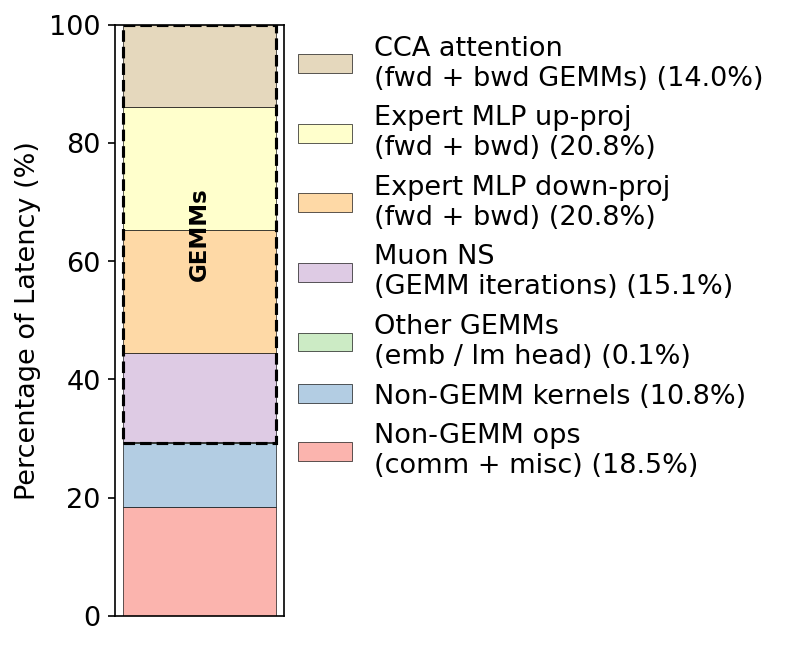}
        \caption*{(b) GPU operation breakdown}
    \end{minipage}
    \caption{Proportion of operations between GEMM kernel compute, non-GEMM kernel compute, and non-compute kernels. Breakdowns are collected by summing and tagging GPU operations from a PyTorch profile.}
    \label{fig:iter-breakdowns}
\end{figure*}

We have broken down the iteration time during core ZAYA1 pretraining (sequence length 4096), and depict the results in Figure~\ref{fig:iter-breakdowns}. Breakdowns are collected by summing and tagging GPU operations from a PyTorch profile. "Optimizer comm" and "Optimizer compute" denote the communication of our distributed Muon implementation and its inherent Newton-Schulz (NS) iterations. While forward and backward computation in the core kernels (i.e. convolutions and softmax attention for CCA, expert MLP GEMMs, etc) takes the majority of iteration time, computation and communication overhead from the distributed Muon optimizer contribute a non-negligible effect. This optimizer compute overhead motivated the Muon kernel in Section \ref{sec:muon-kernel}, and the communicated overhead motivated the tensor fusion tuning described in Sections \ref{sec:pollara-bandwidth} and \ref{sec:infinityfabric-bandwidth}.

\subsection{Kernels}
\label{sec:kernels}

We wrote custom HIP kernels for the LayerNorm operation and for several subroutines used in the Muon optimizer.

\subsubsection{Muon}
\label{sec:muon-kernel}
The Muon optimizer combines Stochastic Gradient Descent with momentum with an added orthogonalization step on the momentum-updated gradient. The approximate orthogonalization is performed via a few iterations of the Newton–Schulz (NS) method on the full momentum-updated gradient matrix. Note that by construction, Muon can only be applied to 2D parameters; other parameters (such as 1D norms and 3D convolutional layers) are optimized with AdamW. Our custom kernels optimize the weight and momentum update steps and the NS step.

Our weight and momentum update kernels are fused HIP multi-tensor kernels, similar to the fused AdamW kernels in Apex \citep{apex}, and are invoked via \texttt{multi\_tensor\_apply}: (i) a \emph{momentum} kernel that updates the FP32 momentum buffer and emits BF16 inputs for the NS method, and (ii) an \emph{update} kernel that applies decoupled weight decay and the update to FP32 master weights. Both kernels process lists of parameter shards in chunks with instruction-level parallelism (ILP) and coalesce memory accesses.

Denote a 2D tensor of trainable parameters as $w$, the associated gradient at step $t$ as $g_t$, and the momentum buffer as $m_t$. The momentum kernel computes the input tensor for the NS procedure:
\begin{align*}
m_t &= \mu\, m_{t-1} + g_t, &
\mathrm{NS_{in}} &=
\begin{cases}
m_t & \text{(no Nesterov)}\\
g_t + \mu\, m_t & \text{(Nesterov)}
\end{cases}
\end{align*}
where $\mu$ is a fixed scalar. The output of the NS procedure is the weight update that is used by the second kernel to compute:
\begin{align*}
w &\leftarrow (1-\eta\,\delta)\,w, &
w &\leftarrow w - \lambda\, \mathrm{NS_{out}}.
\end{align*}
Here, $\eta$ is the base learning rate that is shared with AdamW-optimized parameters, $\delta$ is the weight decay factor and $\lambda$ is the adjusted learning rate for Muon.

All arithmetic accumulates are in FP32; BF16 is used to reduce bandwidth for gradients/NS inputs and optional shadow weights. Separating momentum and Nesterov-update from the weight update (i) produces bandwidth-friendly BF16 NS inputs, (ii) keeps master-weight traffic isolated, and (iii) allows distinct launch/stream tuning for NS vs.\ update. In practice, these two memory-linear passes substantially cut optimizer overhead, such that iteration time is dominated by model compute and communication rather than optimizer step time.

Next we describe the kernel for the NS procedure. The Newton--Schulz iterations in Muon introduce a significant cost when implemented with generic GEMMs, because their core work repeatedly forms a Gram matrix and its powers. Writing the NS step as
\begin{align}
X \leftarrow a\, X + b\, (XX^\top)X + c\, (XX^\top)^2 X,
\end{align}
the expensive pieces are $A \leftarrow XX^\top$ and $A^2 \leftarrow AA$. We replace these with a symmetric matrix multiplication kernel that computes $XX^\top$ (and, by symmetry, $AA = AA^\top$) by tiling the $m\times m$ output, early-exiting tiles strictly below the diagonal, and, in the epilogue, writing both the computed upper tile $(i,j)$ and its transpose to $(j,i)$ back to HBM. Concretely, for $X \in \mathbb{R}^{m\times k}$ the kernel:
\begin{enumerate}
\item assigns one program to each output tile $(i,j)$ in the upper triangle (including the diagonal);
\item streams $k$-sized chunks from rows $i$ and $j$ of $X$, accumulates FMAs in FP32 into a $B_M\times B_M$ register tile;
\item stores the tile to $(i,j)$ and simultaneously stores its transpose to $(j,i)$.
\end{enumerate}
Because $A=XX^\top$ is symmetric, this eliminates roughly half of the multiply--accumulate work and halves HBM writes for off-diagonal tiles (the diagonal is computed once), replacing the saved work with a register-local transpose at store time. The same kernel is reused to form $A^2$ by calling it on $A$ itself, since $A^2=AA^\top$ when $A$ is symmetric. The kernel is stride-aware (no data reorders), and accumulates in FP32 while allowing BF16 I/O to match our momentum pass. We expose both a functional form and an in-place target form to avoid transient allocations inside the NS loop.

Putting this together, one NS step becomes:
\begin{align}
A \leftarrow \mathrm{kernel}(X),\qquad
A^2 \leftarrow \mathrm{kernel}(A),\qquad
X \leftarrow a\, X + (b\, A + c\, A^2)\, X,
\end{align}
with the last multiplication realized as a standard GEMM (row-major friendly) and scalar axpy. In practice, this:
\begin{itemize}[nosep,leftmargin=1.25em]
\item cuts the arithmetic for $A$ and $A^2$ formation by approximately $2\times$ and reduces output-store traffic on off-diagonals by approximately $2\times$;
\item keeps numerical behavior aligned with the reference (differences only from benign FP32 reduction order), since the full $A$ is materialized before use;
\item avoids extra reads/writes.
\end{itemize}
With five NS steps, the symmetric matrix multiplication savings dominate the optimizer's overhead, making Muon's NS phase bandwidth-friendly and shrinking its share of iteration time without altering the algorithm or its stability.

\subsubsection{LayerNorm}
\label{sec:layernorm}

We found that naively applying the Transformer Engine \citep{nvidia_transformerengine} kernel to HIP for layernorm provided subpar performance, motivating us to develop our own optimized kernel. In this kernel, we fuse residual add, statistics, normalization, and affine into a single HIP kernel over $(B\!\cdot\!T)\times E$ rows. For each row with $N\!=\!E$,
\[
v \leftarrow x + \mathrm{residual},\qquad
\mu=\frac{1}{N}\sum_{i=1}^{N} v_i,\qquad
\sigma^2=\frac{1}{N}\sum_{i=1}^{N} (v_i-\mu)^2,\qquad
\widehat v_i=\frac{v_i-\mu}{\sqrt{\sigma^2+\varepsilon}},
\]
\[
y_i=\gamma_i\,\widehat v_i+\beta_i,\qquad
\mathrm{residual\_out}_i=v_i.
\]
Statistics are accumulated in FP32 via single-pass Welford. We store per-row $\mu$ and $\mathrm{inv\_std}=(\sigma^2+\varepsilon)^{-1/2}$ (FP32) for the backward. I/O supports \texttt{bf16}/\texttt{fp32} while compute is \texttt{fp32}. One thread block processes a row; threads stride across $E$. In stage~1 of the kernel, we compute and combine Welford partials in shared memory, while in stage~2, we write $\mathrm{residual\_out}$ and $y$.

Given upstream $g=\partial\mathcal{L}/\partial y$ and saved $(\mu,\mathrm{inv\_std})$, the backward forms
\[
S_1=\sum_{i=1}^{N} g_i\gamma_i,\qquad
S_2=\sum_{i=1}^{N} (g_i\gamma_i)\,\widehat v_i,
\]
and computes
\[
\frac{\partial\mathcal{L}}{\partial v_i}
=\frac{\mathrm{inv\_std}}{N}\Big(N\,g_i\gamma_i - S_1 - \widehat v_i\,S_2\Big).
\]
Since $v=x+\mathrm{residual}$,
\[
\frac{\partial\mathcal{L}}{\partial x_i}
= \frac{\partial\mathcal{L}}{\partial \mathrm{residual}_i}
= \frac{\partial\mathcal{L}}{\partial v_i}.
\]
Parameter gradients are standard reductions over rows $t$:
\[
\frac{\partial\mathcal{L}}{\partial \gamma_i}=\sum_{t} g_{t,i}\,\widehat v_{t,i},
\qquad
\frac{\partial\mathcal{L}}{\partial \beta_i}=\sum_{t} g_{t,i}.
\]
We also implement an RMSnorm variant by dropping the mean subtraction and using
\[
\widehat v_i^{\mathrm{RMS}}=\frac{v_i}{\sqrt{\frac{1}{N}\sum_{j=1}^{N} v_j^2+\varepsilon}},
\]
with the same fusion and reduction structure.

\subsection{Parallelism}
\label{sec:parallelism}

\subsubsection{Context-Parallelism}
\label{sec:context-parallelism}
In order to extend the context of ZAYA1 base, we use context-parallelism to shard the sequence among multiple GPUs. Specifically, we shard the sequence into $2\times(\textrm{context-parallel world size})$ GPUs where each GPU holds $(\textrm{sequence length})/(2\times(\textrm{context-parallel world size}))$ tokens. These shards are shuffled in order to maintain load balancing during the causal attention computation. We use the context-parallel attention algorithm known as Ring Attention \citep{liu2023ring} which sends key and value shards in a point to point manner between context-parallel ranks that form a logical ring pattern. Local attention is computed and renormalized as each rank receives new key and value shards. This computation is overlapped with communication. 

\begin{figure*}[htbp]
    \centering
    \includegraphics[width=0.75\linewidth]{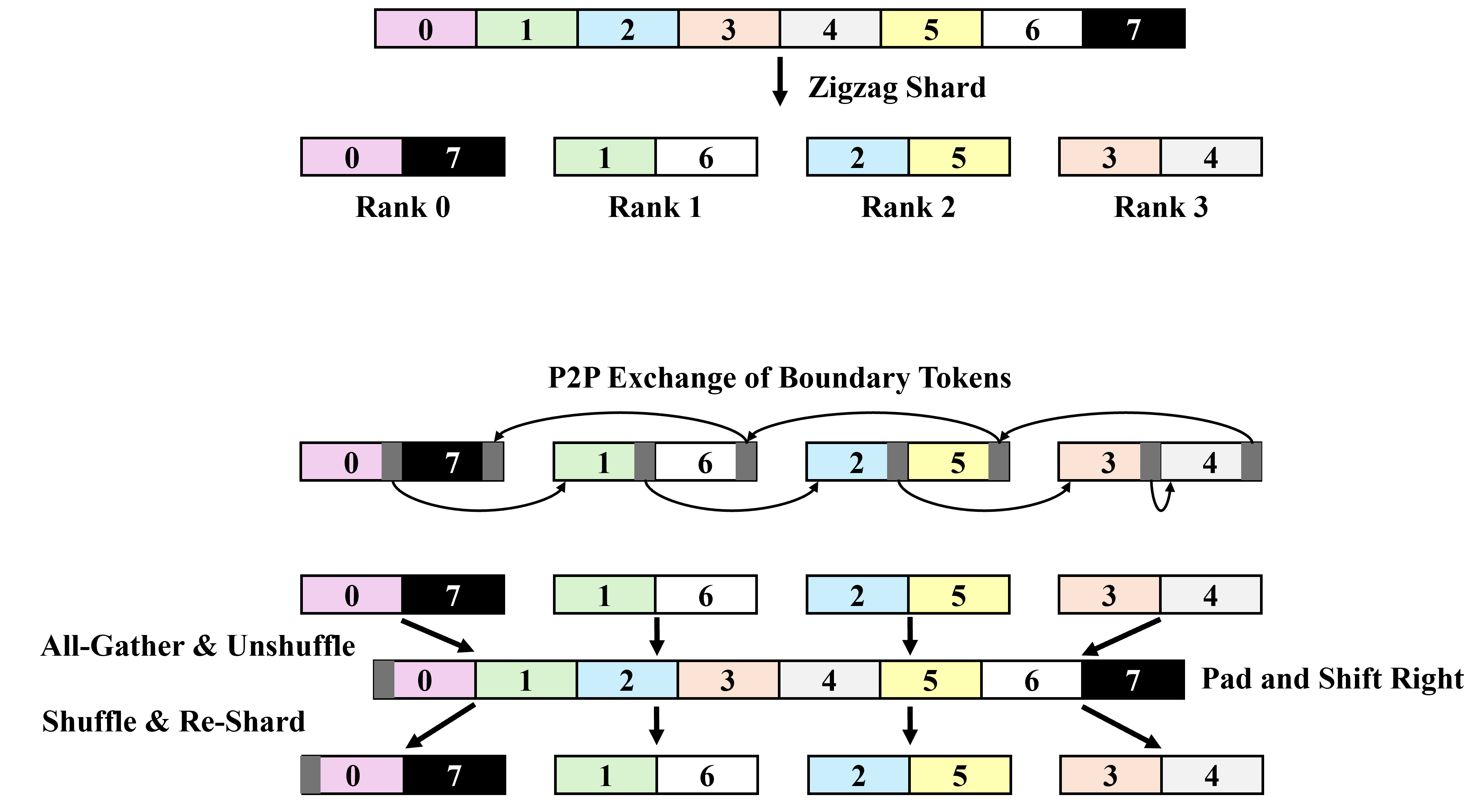}
    \caption{The context-parallelism design used to train ZAYA1-base on longer context lengths.}
    \label{fig:cp-design}
\end{figure*}

We needed to develop a novel communication scheme for the parallelization of the $qkv$ preprocessing that CCA requires (see appendix \ref{app:cca}). In particular, the convolutions and value-shift are new forms of sequence mixing that need to be accounted for when sharding the context. Given that the two sequential convolutions we employ have a very small width (2 each), we just send the final two tokens from the end of each of the sequence chunks to the rank holding the subsequent chunk. Note that because of the load balanced sharding pattern, we send token chunks bidirectionally in a manner that is depicted in Figure~\ref{fig:cp-design}. After this low volume communication, every rank can simultaneously perform the convolution operations locally, and the end result will be the appropriately sharded result of performing the convolutions of the entire sequence. Then for the value shift and padding, we simply perform the projection by the second value parameter, which compresses the hidden dimension of the embeddings down to a single head’s worth (because we have just 2 value heads and the second value projection accounts for half of these heads). In the compressed space, we can easily perform an all gather along the sequence axis, and then pad the beginning of the gathered sequence and shift it right by one position, and then re-shard the result. 

In the backward pass, for the point to point exchange of chunk-boundary tokens, we need only replace every send operation with a receive operation and vice versa. Then we can simultaneously compute the vector Jacobian products for the two convolution operations on each rank. For the value projection, we replace the AllGather with a ReduceScatter in the backward pass and differentiate through the projection. In both forward and backward passes we ensure that the communication overhead remains minimal.

\subsubsection{Muon}

\begin{figure}
    \centering
    \includegraphics[width=0.75\linewidth]{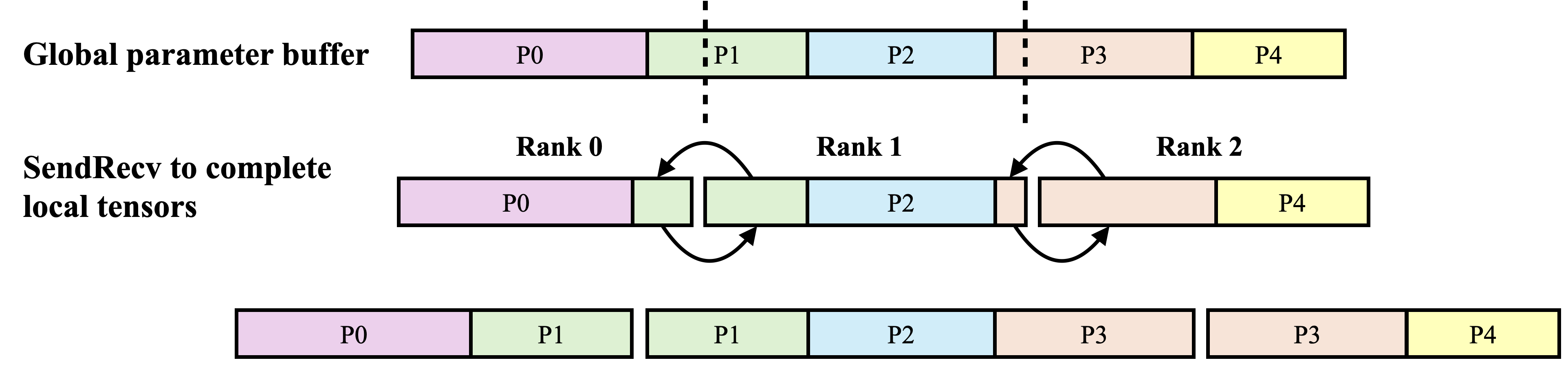}
    \caption{The \texttt{SendRecv} communication we implemented for the Muon step to keep the memory overhead low based on the parameter sharding scheme.}
    \label{fig:ddp-sharding}
\end{figure}

Because Muon’s update step involves orthogonalizing the gradient matrix, it cannot operate element-wise like AdamW. Under the ZeRO-1 setup that we used for training, parameters are sharded across data-parallel (DP) ranks. Consequently, during each optimizer step, each optimizer instance must assemble the full gradient matrix for each parameter that it owns. The required communication pattern depends on the parameter sharding scheme. In our case, parameter tensors are flattened, concatenated, and then divided into contiguous chunks (without respecting parameter boundaries), which are distributed across DP ranks, as illustrated in Fig.~\ref{fig:ddp-sharding}. An alternative sharding scheme splits each parameter tensor into as many chunks as there are DP ranks, so that every rank holds a slice of every parameter. This scheme is less favorable for Muon, because it increases the communication load.

As a base, we used Kimi's Megatron-LM implementation of Muon~\citep{megatron_distributedmuon}, where the distributed Muon step uses a simple but memory-intensive procedure: parameters are written into a global buffer via an \texttt{AllGather}. Then, each optimizer instance performs the Newton–Schulz iterations on the parameters it owns, extracts the required slices, and applies the weight updates to its local shards. This causes a significant temporary memory spike, leading to a higher peak memory\footnote{When considering parallelism schemes, the peak memory is the primary constraint one seeks to reduce, as it determines whether your device is able to complete a given training iteration.} and largely negates the benefit of ZeRO-1; for ZAYA1, the memory overhead incurred by \texttt{AllGather} is $99\%$ of the total optimizer memory. However, given our sharding scheme, the \texttt{AllGather} is unnecessary because parameters are only split at rank boundaries, i.e., most parameters assigned to a rank remain whole, and at most two (those at the shard edges) are split. We therefore replace the \texttt{AllGather} with a \texttt{SendRecv} exchange, where each rank communicates only the incomplete parameter shards with its immediate neighbors. Specifically, for each parameter in the optimizer step: if the parameter is wholly contained within a rank's shard, it is reshaped and used directly; if only one part is available locally, the rank exchanges its incomplete portion with the neighboring rank via point-to-point communication (send to neighbor, receive from the same neighbor), then concatenates the received portion with its local data to reconstruct the full parameter tensor. The ordering of send and receive operations is determined by rank comparison to avoid deadlock: the lower-ranked process sends first, then receives; the higher-ranked process receives first, then sends.

\begin{algorithm}[ht]
\caption{Memory-efficient distributed Muon with \texttt{SendRecv}.}
\begin{algorithmic}[1]
\For{each parameter $p$ in optimizer shard}
    \If{$p$ is complete}
        \State reshape $p$ into its 2D shape and proceed to Newton-Schulz iterations
    \Else
        \State $\text{neighbor\_rank} \gets$ previous rank if $p$ at shard start, else next rank
        \If{$\text{this\_rank} < \text{neighbor\_rank}$}
            \State \texttt{send}(local portion of $p$, neighbor\_rank)
            \State \texttt{recv}(missing portion of $p$, neighbor\_rank)
        \Else
            \State \texttt{recv}(missing portion of $p$, neighbor\_rank)
            \State \texttt{send}(local portion of $p$, neighbor\_rank)
        \EndIf
        \State Concatenate received and local portions to form complete $p$, reshape to 2D, proceed to Newton-Schulz iterations
    \EndIf
\EndFor
\end{algorithmic}
\end{algorithm}

\subsection{Checkpointing}
\label{sec:checkpointing}

One underappreciated point in the training stack is checkpointing. Pretraining at scale requires checkpointing that is fast, reshape-friendly, and robust to routine hardware and network faults. The default centralized approach of gathering optimizer state is not scalable and creates a single point of failure. We therefore adopt distributed checkpointing in which each rank persists its own optimizer shard and lightweight metadata, while model weights are written once by the root rank. This is effective because the optimizer states are by far the largest part of the checkpoint. The device-to-host staging is a blocking copy to ensure data integrity and consistency. The copy is parallelized using a threadpool to accelerate the operation. We then perform the final write to disk as an asynchronous process, overlapped with post-checkpoint training iterations in order to resume training as quickly as possible after the checkpoint save. Checkpoints are marked complete only after all artifacts finish writing, and the start of the next checkpoint cycle waits for any remaining asynchronous processes to complete, which are subsequently reaped. In practice, this removes the root-rank I/O bottleneck and reduces visible pause time, while a moderate checkpoint cadence avoids contention with background writers of the previous checkpoint. In practice, we find that our checkpointing scheme reduces checkpoint time by more than 10$\times$ compared to naive checkpointing, which then allows us to increase the checkpoint frequency, reducing the time-cost of failures and restarts.

Our training employs Muon for specific parameter subsets and AdamW for the remainder. AdamW optimizes the CCA convolutions (both 1D convs), tied embeddings, residual-scaling vectors, all layer norms (RMSNorms, excluding qk-norm), and CCA temperatures. All remaining parameters—Q/K/V projections, output projections, router MLPs, down-projections, and MoE experts—are optimized with Muon. We model the distributed checkpointing procedure as follows: Let
\(P\) be the total number of trainable parameters saved, with $P=P_M+P_A$, where $P_M$ and $P_A$ are the numbers of parameters optimized with Muon and AdamW, respectively. Let 
\(b_{\mathrm{lp}}\) be the bytes per low-precision weight,
\(b_{\mathrm{hp}}\) the bytes per high-precision value (\(b_{\mathrm{lp}}=2\) for \texttt{bf16} and \(b_{\mathrm{hp}}=4\) for \texttt{fp32}),
\(\texttt{dp\_degree}\) the ZeRO-1 data-parallel degree,
and \(m_r\) a small per-rank metadata memory overhead in bytes (RNG states, schedulers, JSON, etc.). 

The optimizer state to store per parameter optimized with Muon is
\[
s_{\text{Muon}}
\;=\;
\underbrace{b_{\mathrm{hp}}}_{\text{master copy}}
\;+\;
\underbrace{b_{\mathrm{hp}}}_{\text{momentum}}
\;=\;
2 \cdot b_{\mathrm{hp}}
\quad \text{bytes/parameter}
\]
while for parameters optimized with AdamW two buffers are maintained:
\[
s_{\text{AdamW}}
\;=\;
\underbrace{b_{\mathrm{hp}}}_{\text{master copy}}
\;+\;
\underbrace{2\cdot b_{\mathrm{hp}}}_{\text{1st and 2nd moment estimates}}
\;=\;
3 \cdot b_{\mathrm{hp}}
\quad \text{bytes/parameter}
\]
The total checkpoint size written to (or read from) disk is therefore:
\begin{align}
S_{\mathrm{total}}
\;=\;
\underbrace{P \cdot b_{\mathrm{lp}}}_{\text{low-precision weights}}
\;+\;
\underbrace{P_M \cdot 2 \cdot b_{\mathrm{hp}}}_{\text{Muon optimizer (all ranks)}}
\;+\;
\underbrace{P_A \cdot 3 \cdot b_{\mathrm{hp}}}_{\text{AdamW optimizer (all ranks)}}
\;+\;
\sum_{r=1}^{\texttt{dp\_degree}} m_r
\quad \text{bytes}.
\end{align}

Since the optimizer states dominate the checkpoint size and are distributed across \(\texttt{dp\_degree}\) (ZeRO-1), we decrease save time by having each data-parallel rank write its own optimizer shard in a separate file. The root rank additionally writes the consolidated weights:
\begin{align}
S_{\mathrm{rank\_0}}
&=
\underbrace{P \cdot b_{\mathrm{lp}}}_{\text{weights}}
\;+\;
\underbrace{\frac{(2 P_M + 3 P_A) \cdot b_{\mathrm{hp}}}{\texttt{dp\_degree}}}_{\text{optimizer state shard}}
\;+\; m_0
\quad \text{bytes},\\[4pt]
S_{\mathrm{rank\_r}}
&=
\underbrace{\frac{(2 P_M + 3 P_A) \cdot b_{\mathrm{hp}}}{\texttt{dp\_degree}}}_{\text{optimizer state shard}}
\;+\; m_r
\qquad (r\neq 0)
\quad \text{bytes}.
\end{align}

Resuming training with a different world size is handled via offline reshaping of optimizer shards. Our training framework's padding scheme (rounding shard sizes such that the global parameter vector is divisible by alignment and GPU count) enables a deterministic unpad–remap–repad procedure. 
Combined with our fault-tolerance service \texttt{Aegis} (see Section~\ref{sec:fault-tolerance}), this scheme provides fast and flexible recovery throughout long runs. Being able to reshape checkpoints on the fly enables recovery even with arbitrary node failures, at the cost of minor automatic adjustments to the global batch size. We found this ability very helpful in practice in a variety of circumstances.

\subsection{Storage and I/O}
\label{sec:storage-node}

Our storage needs are met by a single storage node whose hardware is further described in Section \ref{sec:cluster-setup}. The two main loads on the storage node are checkpointing and dataloading. We performed detailed calculations to ensure that the storage node's capacity and bandwidth would be sufficient for Zyphra's training needs. Some of these calculations are provided in Appendix \ref{app:storage-io}. Training small models requires high IOPS since iteration times are quick and GPUs demand data faster from the storage node. Simply adding more CPU dataloader workers only delays GPU starvation in this case, which is evidenced in training as a sudden increase in iteration time: $iter\_time_{baseline}=Time(fwd+bwd)\rightarrow Time(load+fwd+bwd)$. Specifically, dataloader workers fetch data in advance before the first iteration is reached (during setup and checkpoint loading), which hides the dataloading bottleneck temporarily, but since $time(iter) < time(dataloader\_thread)$ we reach the bottlenecked steady-state over time. Therefore, we must decrease the proportion of scattered reads by producing \emph{sequential data shards}\footnote{Most LLM training dataloaders pull from random documents that comprise the component datasets, which leads to many scattered reads that stress the storage hardware (see Appendix \ref{app:storage-io}). We instead prepackage a shard that contains future samples, and overlap shard creation on idle nodes' RAM with training iterations on the previous shard. Once the old shard is exhausted, we load in the new shard}, non-blocking dataloading, and storage hardware powerful enough to support the bandwidth needed by smaller models.

While smaller models stress the storage hardware IOPS, large models stress the bandwidth due to their massive checkpoints. We leave the discussion of our designs to subsequent papers discussing larger models, but steps were made to mitigate checkpoint saves (see Section \ref{sec:checkpointing}) and loads. Specifically for loads, we greatly increased the size of each nodes' page cache to 1\,TB of the total available RAM. Since the page cache lives in the OS and not the training processes, it persists after program termination, and subsequent loads of the checkpoint are nearly instant from RAM instead of from local NVMe or the storage node. Since text training sequences are comparatively small, we checkpoint often, and our page cache is so large, we're able to keep several of the most recent checkpoints in RAM. This is helpful if we need to rewind (loss spikes due to HBM corruption, divergences, etc). We also note that this allows for the training samples within the overlap between $rewind\_iter$ and $max\_reached\_iter$ to be resident in cache and immediately loaded until training reaches an iteration count higher than seen beforehand. This design, in addition to two-level checkpointing designs inspired by classical HPC applications \citep{scr, scr-dl}, allows checkpoint loads and stores to minimize stress on the filesystem hardware and reduce application downtime.

\subsection{Fault Tolerance}
\label{sec:fault-tolerance}

\begin{figure*}[htbp]
    \centering
    \includegraphics[width=0.75\linewidth]{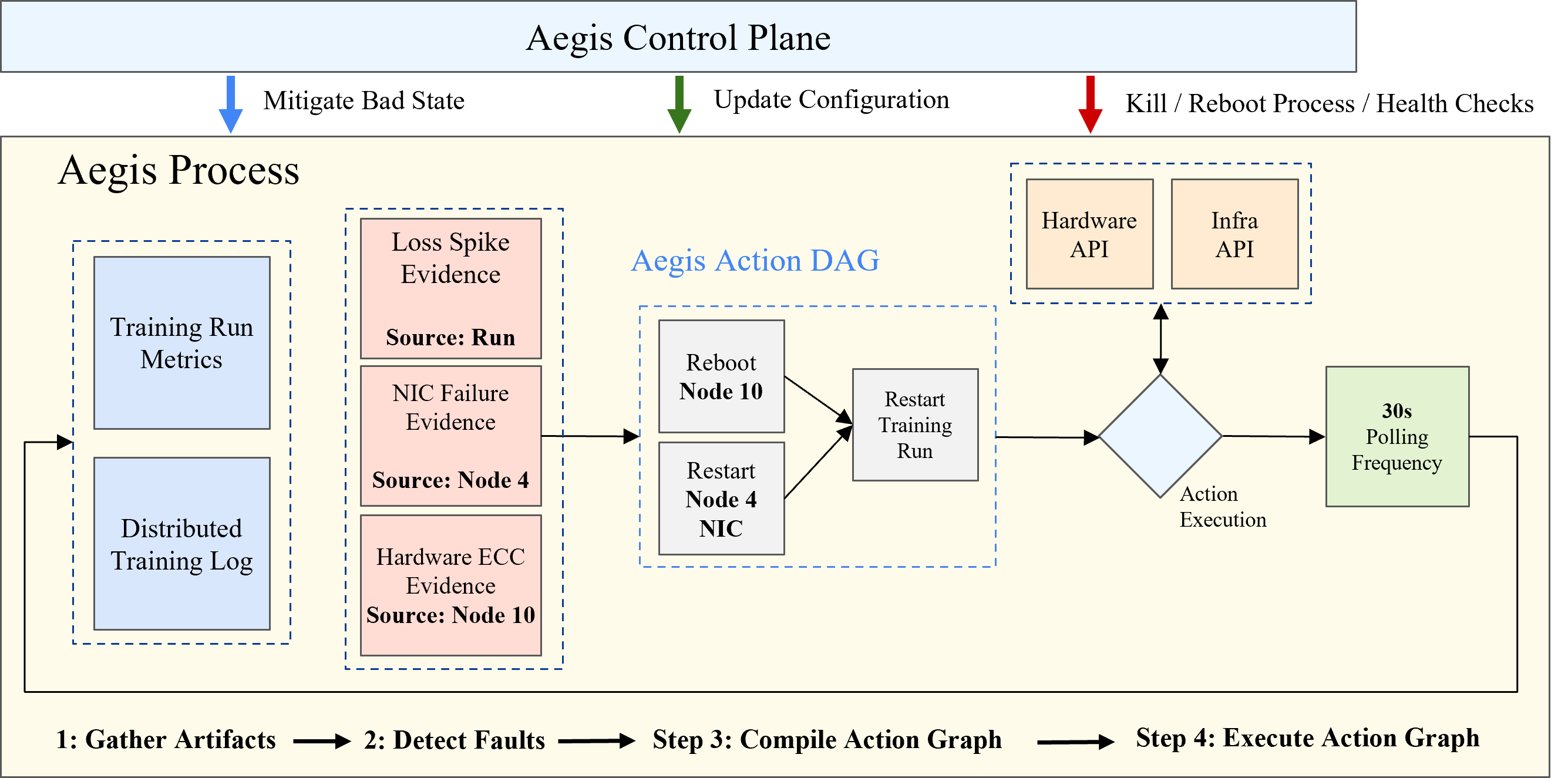}
    \caption{High-level architecture of the Aegis fault tolerance system. 
    The system operates as a four-step process consisting of (1) artifact collection, 
    (2) evidence gathering, (3) action graph compilation, and (4) action graph execution. 
    Additionally, we implement a control plane that allows human triage efforts to 
    dynamically scope Aegis's response and take appropriate actions to refresh failing 
    Aegis processes.}
    \label{fig:aegis-diagram}
\end{figure*}

Additionally, we leverage an in-house fault tolerance system that we call \textit{Aegis}, which we have developed to help minimize downtime in the event of hardware, networking, or other training failure. Broadly speaking, this system polls training artifacts such as logs and run observability platforms at a constant frequency, identifies when run failures appear, and then notifies us of the failure and its likely cause. In the case of well-known situations with a clear-cut fix (e.g. network failures, loss spikes caused by data corruption, individual node failures), the system is equipped with capabilities to directly and automatically take action and a channel to clearly telegraph the action taken to coordinate with human triage responses. The Aegis system is connected directly to the IBM cloud CLI and platform, enabling it to take actions such as creating, rebooting, or deleting failed nodes automatically. For a system diagram of the Aegis system, see Figure~\ref{fig:aegis-diagram}. 

Below we list certain common types of errors that we have encountered and their mitigation.

\begin{itemize}
    \item \textbf{Completion Queue Errors (CQE):} Many hardware issues with the AMD Pollara interconnect manifest as CQE errors, which can require either a restart of the Pollara NICs, restarting the port, or physically replacing/cleaning the transceiver. On large-scale clusters, these errors can be tedious to debug as they require running communication RCCL tests to identify faulty NICs. We have developed a series of scripts that can be plugged into Aegis to automatically detect faulty NIC cards, track down failures, and mitigate several common NIC misconfiguration issues.
    \item \textbf{Generic RCCL Communication Errors:} These errors typically occur due to malfunctions within the network interface card (NIC) or the communication middleware running atop it (RCCL or PyTorch's RCCL interface). These errors are treated similarly to CQE errors and often manifest when the network is down either due to node or NIC failure. 
    \item \textbf{GPU ECC or GPU hang error:} These errors occur when there is some kind of hardware failure either of the DRAM or inside the GPU itself. If detected, our system automatically selects a new node and restarts the run, while sending the node back to IBM for repair.
    \item \textbf{Flapping at switches or NICs:} One common issue we regularly encounter is that a NIC runs into some error during transmission or reception of data and 'flaps' causing the link to reset itself. Such events typically result in overall run hangs of 60-200 second duration and will cause a timeout and failure of the run if handled incorrectly. We adapted to the general frequency of these events by setting long RCCL timeouts, which prevents crashes at the cost of potentially un-noticed hangs.
\end{itemize}

Also potentially of interest is that we experienced occasional sudden loss spikes which we traced back to silent data corruption events occurring on a single faulty GPU. Prior to tracing this node these events would occur approximately once per week. These were handled by automatically restarting the run when such a loss spike was detected which would cause the run to resume its pre-spike trajectory. 

Generally, our philosophy with the Aegis system is that we handle automatically all known failure cases with straightforward `rote' resolution. When new failures are detected, these still require human oversight and intervention for now, however each new failure case adds to our decision tree of actions to be taken to restore operation as soon as possible. In the longer term, we want to develop Aegis into a system which can handle small events such as individual node or NIC failures without even requiring training restart but instead can seamlessly swap in and out new nodes while the run is still ongoing in order that we can approach 100\% uptime even with nontrivial rates of hardware failure. We envision this capability becoming increasingly important as we scale to larger clusters with greater expected failure rates.


\section{Results}
\label{sec:results}

Here we present preliminary benchmark results for our base model, comparing against representative models with parameter counts between 1B and 8B. ZAYA1-base has approximately 760M active parameters and 8.3B total parameters. We find that it roughly performs around the level of a very strong 4B dense model, for instance, it performs competitively with Qwen3-4B \citep{yang2025qwen3} despite having more than 4 times fewer active parameters. ZAYA1-base is also surprisingly competitive with 7-12B base models of the prior generation, despite them having more active parameters than ZAYA1-base has total. ZAYA1-base comfortably outperforms Llama3-8B \citep{llama3} across the board while exceeding even Gemma3-12B \citep{team2025gemma} in challenging mathematics and coding benchmarks. 


\begin{figure}
     \centering
     \includegraphics[width=0.75\linewidth]{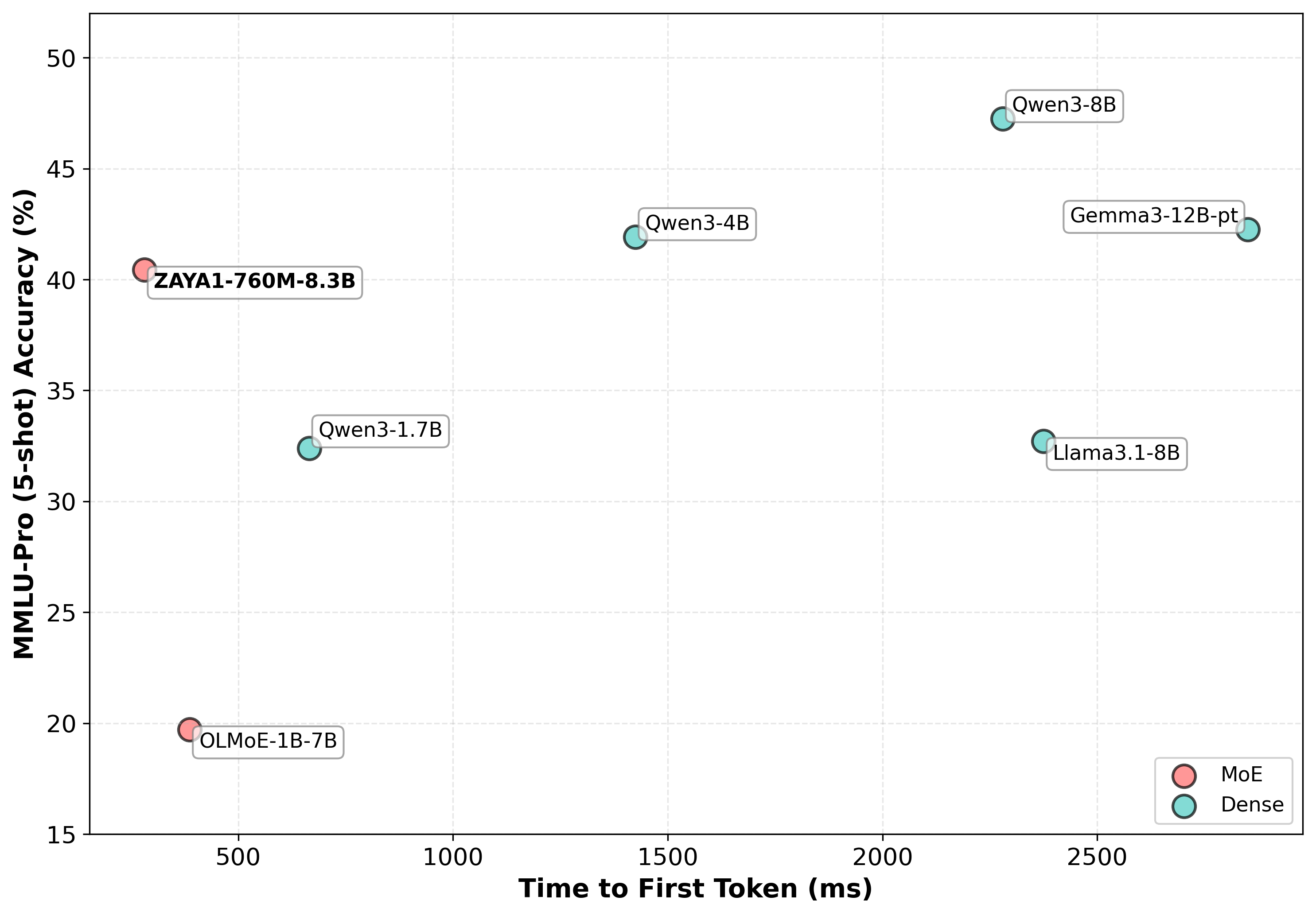}
     \caption{Performance of ZAYA1-base vs comparable base models at different scales of time to first token (TTFT) and advanced general knowledge capability (MMLU-pro).}
     \label{fig:mmlu-pro}
\end{figure}

In terms of MoE models, ZAYA1-base outperforms the recently released MoE models of similar scales on mathematics and general knowledge evaluations, while lagging slightly in coding. ZAYA1-base dramatically outperforms prior open MoE models such as OLMoE \citep{muennighoff2024olmoe} demonstrating both to the strength of the ZAYA1 architecture as well as to improvements in broader pretraining recipes and datasets that have occurred recently. 

\begin{table}[ht]
\centering
\small  
\setlength{\tabcolsep}{2pt}
\renewcommand{\arraystretch}{1.12}

\begin{tabular}{lccccc}
\toprule
\textbf{Model} & \textbf{MMLU(0)} & \textbf{MMLU-Pro(5)} & \textbf{GPQA(0)} & \textbf{MATH-hard(4)} & \textbf{MBPP+(3)} \\
& Acc(\%) & Acc(\%) & Acc(\%) & Exact-Match(\%) & Pass@1(\%) \\
ZAYA1-base & 67.01 & 40.43 & 30.70 & 54.15 & 75.40 \\
\href{https://huggingface.co/Qwen/Qwen3-1.7B-Base}{Qwen3-1.7B} 
& 54.10&	32.4 &	28.9 &	33.2 &	55.82 \\
\href{https://huggingface.co/Qwen/Qwen3-4B-Base}{Qwen3-4B}
& 68.31 & 41.92	& 33.72	& 47.05	& 76.46 \\
\href{https://huggingface.co/allenai/OLMoE-1B-7B-0125}{OLMoE-1b-7b} & 53.40 & 19.71 & 26.34 & 4.68 & 29.10 \\
\href{https://huggingface.co/Qwen/Qwen3-8B-Base}{Qwen3-8b} 
& 74.62	& 47.24	& 36.07	& 28.17	& 81.00 \\
\href{https://huggingface.co/google/gemma-3-12b-pt}{Gemma3-12b-pt} & 71.44 & 42.25 & 35.15 & 17.98 & 75.40 \\
\href{https://huggingface.co/meta-llama/Llama-3.1-8B}{Llama3.1-8b} & 63.29 & 32.71 & 31.12 & 6.50 & 62.70 \\
\bottomrule
\end{tabular}

\caption{Comparison of ZAYA1-base on core general knowledge, mathematics, and coding evaluations. 
ZAYA1-base performs extremely strongly considering its active parameter count, significantly outperforming similar MoE models such as OLMoE and going head to head against extremely strong much larger dense models such as Qwen3-4b and Gemma3-12b.}
\label{tab:lm-eval-scores}
\end{table}

ZAYA1's reasoning-focused mid-training checkpoint shows very strong pass@k performance, outperforming models such as DeepSeek-R1-Distill-Qwen-7B and approaching SOTA level dense models such as Qwen3-4B even before SFT and RL, indicating high potential for further post-training.

\newcolumntype{Y}{>{\centering\arraybackslash}X}   
\newcolumntype{G}{@{\hspace{8pt}}}                 

\begin{table}[!htbp]
\centering
\caption{Performance of ZAYA1-base vs reasoning models at mathematics and general knowledge evaluations using chain of thought reasoning. We measure performance at high pass@k (best of 64) in order to get a preliminary sense of the model's strength once post-training is applied. We see that at high pass@k, ZAYA1-base performs extremely strongly against full-fledged reasoning models such as Phi-4-Mini-Reasoning and Deepseek-R1-7B, nearly matching state-of-the-art reasoning models at this scale such as the latest Qwen3-4B-Thinking. This is despite ZAYA1-base not having yet undergone any instruct or RL post-training.}
\label{tab:comprehensive-eval}
\small
\setlength{\tabcolsep}{2pt}
\renewcommand{\arraystretch}{1.12}
\begin{minipage}{\linewidth}\centering

\begin{tabularx}{\linewidth}{l*{7}{Y}G Y Y G Y Y}
\toprule
& \multicolumn{7}{c}{\textbf{MathArena best@64}}
  & \multicolumn{2}{c}{\textbf{LCB best@16}}
  & \multicolumn{2}{c}{\textbf{MCQA best@16}} \\
\cmidrule(lr){2-8}\cmidrule(lr){9-10}\cmidrule(lr){11-12}
\textbf{Model} & \textbf{aime24} & \textbf{aime25} & \textbf{amc23} & \textbf{apex} & \textbf{brumo} & \textbf{hmmt} & \textbf{cmimc} & \textbf{v5} & \textbf{v6} & \textbf{gpqa\_d} & \textbf{mmlu\_pro} \\
\midrule
ZAYA1-reasoning-base         & 87.90\% & 84.50\% & 100.00\% & 17.02\% & 91.66\% & 79.16\% & 76.26\% & 74.40\% & 66.04\% & 91.18\% & 91.56\% \\
\href{https://huggingface.co/Qwen/Qwen3-4B-Thinking-2507}{Qwen3-4B-Thinking-2507}      & 92.16\% & 91.95\% & 100.00\% & 6.97\%  & 93.33\% & 82.69\% & 78.19\% & 70.44\% & 69.21\% & 84.58\% & 86.13\% \\
\href{https://huggingface.co/Qwen/Qwen3-4B}{Qwen3-4B}                    & 87.56\% & 88.71\% & 100.00\% & 16.78\% & 91.59\% & 80.45\% & 68.65\% & 63.68\% & 59.50\% & 77.49\% & 82.76\% \\
\href{https://huggingface.co/microsoft/Phi-4-mini-reasoning}{Phi-4-Mini-Reasoning}        & 84.11\% & 76.03\% & 99.38\%  & 11.00\% & 77.57\% & 60.40\% & 66.13\% & 44.14\% & 42.27\% & 86.76\% & 86.08\% \\
\href{https://huggingface.co/HuggingFaceTB/SmolLM3-3B}{SmolLM3-3B}                  & 81.92\% & 82.30\% & 97.38\%  & 11.11\% & 83.87\% & 64.38\% & 67.16\% & 52.39\% & 47.91\% & 84.56\% & 91.66\% \\
\href{https://huggingface.co/deepseek-ai/DeepSeek-R1-Distill-Qwen-7B}{DeepSeek-R1-Distill-Qwen-7B} & 86.02\% & 79.00\% & 99.07\%  & 11.10\% & 85.27\% & 63.08\% & 71.50\% & 56.22\% & 49.56\% & 85.02\% & 85.76\% \\
\bottomrule
\end{tabularx}

\vspace{6pt}

\begin{tabularx}{\linewidth}{l*{7}{Y}G Y Y G Y Y}
\toprule
& \multicolumn{7}{c}{\textbf{MathArena avg@64}}
  & \multicolumn{2}{c}{\textbf{LCB avg@16}}
  & \multicolumn{2}{c}{\textbf{MCQA avg@16}} \\
\cmidrule(lr){2-8}\cmidrule(lr){9-10}\cmidrule(lr){11-12}
\textbf{Model} & \textbf{aime24} & \textbf{aime25} & \textbf{amc23} & \textbf{apex} & \textbf{brumo} & \textbf{hmmt} & \textbf{cmimc} & \textbf{v5} & \textbf{v6} & \textbf{gpqa\_d} & \textbf{mmlu\_pro} \\
\midrule
ZAYA1-reasoning-base         & 62.39\% & 54.26\% & 91.13\% & 0.52\% & 61.67\% & 32.03\% & 34.75\% & 52.76\% & 45.75\% & 53.36\% & 68.48\% \\
\href{https://huggingface.co/Qwen/Qwen3-4B-Thinking-2507}{Qwen3-4B-Thinking-2507}      & 75.21\% & 72.60\% & 99.65\% & 0.17\% & 73.44\% & 50.78\% & 49.64\% & 58.98\% & 53.24\% & 66.69\% & 78.93\% \\
\href{https://huggingface.co/Qwen/Qwen3-4B}{Qwen3-4B}                    & 71.98\% & 62.14\% & 96.72\% & 0.52\% & 66.88\% & 42.85\% & 41.20\% & 50.67\% & 44.94\% & 54.35\% & 72.23\% \\
\href{https://huggingface.co/microsoft/Phi-4-mini-reasoning}{Phi-4-Mini-Reasoning}        & 48.39\% & 33.23\% & 85.78\% & 0.69\% & 45.99\% & 21.56\% & 23.96\% & 25.52\% & 23.66\% & 46.67\% & 67.50\% \\
\href{https://huggingface.co/HuggingFaceTB/SmolLM3-3B}{SmolLM3-3B}                  & 46.20\% & 37.40\% & 86.72\% & 2.26\% & 55.16\% & 23.75\% & 25.99\% & 29.91\% & 27.00\% & 43.46\% & 67.05\% \\
\href{https://huggingface.co/deepseek-ai/DeepSeek-R1-Distill-Qwen-7B}{DeepSeek-R1-Distill-Qwen-7B} & 53.49\% & 40.26\% & 89.45\% & 1.04\% & 54.01\% & 27.38\% & 23.33\% & 36.25\% & 32.92\% & 47.94\% & 57.05\% \\
\bottomrule
\end{tabularx}

\end{minipage}
\end{table}

Overall, while the benchmark results we present show that ZAYA1-base is competitive against many leading models, it is still ultimately a base model, and in future work we are excited to see how its capabilities can be developed and enhanced through targeted instruction following and reasoning post-training phases. Our pass@64 results are very encouraging that the base model is strong enough to enable conversion to an exceptionally strong reasoning model during post-training. 

\section{Discussion}

In this paper, we have documented and described in substantial detail our experiences pretraining on a large-scale end-to-end AMD cluster including both MI300x GPUs and Pensando Pollara networking cards including detailed cluster benchmarking and scaling studies. We have also described in detail the optimizations and design choices that go into making an architecture efficient for training and inference on a specific device, as well as given some details on the training framework that we developed to train our core model. Perhaps the key takeaway should be that the AMD hardware, software, and network stack are sufficiently mature and robust to enable large scale LLM pretraining. While the conversion of our training stack to AMD required a small amount of manual work and code conversions, especially relating to porting several kernels required for high performance training, broadly the process was straightforward and a bare-bones version of Megatron mostly worked out of the box, demonstrating the maturity of the ROCm ecosystem and its PyTorch integrations.

Our ZAYA1-base model incorporates a novel architectural recipe consisting of CCA, the ZAYA1 router, and residual scaling. We find that this architecture provides considerable improvements to contemporary MoE architectures measured in loss per FLOP and per parameter. Specifically, CCA allows for large reductions in prefill compute and memory costs, and hence training time, as well as improving loss and making long context training less demanding, while matching the decode speed of existing attention methods such as MLA. Meanwhile the residual scaling and the ZAYA1 router significantly improve the loss per parameter of the model and outperform in FLOP and parameter matched baselines. The ZAYA1 router especially unlocks the latent capacity of the model through superior expert choices and enabling greater expert specialization. We find that utilizing a more powerful and expressive router makes utilizing high top-Ks less important and eliminates the need for residual experts, leading to our architectural choices diverging from the rapidly emerging standard of extremely finegrained experts with high top-k. Instead, we achieve higher sparsity through top-1 with larger experts but more certain routing decisions. Overall, we believe high quality and efficiency of ZAYA1 should make it a powerful model for low-end consumer GPUs as well as for on-device inference. 

In this paper, we have focused primarily on the pretraining on AMD and the pretrained base model. The full potential of the model, however, will only be visible after full post-training especially for RL via reasoning. We plan to pursue these avenues in the future. 

We have also discussed our pretraining stack in detail, especially focusing on the kernels and optimization methods underlying efficient training as well as our fault tolerance and robustness system, topics which are under-discussed in similar papers but are vitally important to training efficiently and stably in practice. Additionally, following recent works \citep{team2025kimi,liu2025muon}, we also utilized the Muon optimizer during pretraining. In our preliminary ablations, we also observed that it outperformed AdamW in longer training runs although it required higher learning rates and larger batch sizes. We found that Muon only added a relatively small overhead to our optimizer step cost, despite using 5 Newton-Schulz iterations, since we were able to improve its efficiency through our kernel work. We also found that selecting the optimal learning rate for Muon was sensitive to the critical batch size being used, as well as that the question of batch size becomes increasingly important for MoE models especially as the degree of sparsity increases.

\section{Conclusion}

This work presents the first comprehensive case study of large-scale language model pretraining on AMD infrastructure, demonstrating that the MI300X GPU and Pollara networking stack are production-ready for frontier-scale training. Our contributions span systems characterization and practical training infrastructure. We provide the first detailed networking benchmarks for Pollara across all major collectives at scale, establish MI300X-specific transformer sizing guidelines, clarify memory bandwidth characteristics, and document our complete cluster architecture in detail. On the software side, we detail our fault-tolerance system (Aegis), checkpoint reshaping utilities, context-parallelism design for CCA, and custom kernel implementations—including fused optimizer operations, layer normalization, and matrix-transpose kernels—that collectively enable competitive throughput.

The ZAYA1-base model validates our architectural innovations: CCA dramatically reduces both prefill compute and KV-cache requirements; the ZAYA1 router enables superior expert specialization; and lightweight residual scaling provides fine-grained information flow control. Training to 12T tokens across three phases required addressing numerous hardware-specific challenges, all of which we document to accelerate future AMD-based efforts. Our results confirm that the AMD ecosystem has matured sufficiently to represent a viable alternative for large-scale LLM development.

\clearpage

\section{Acknowledgements}

This paper would not have been possible without deep collaborations with both AMD and IBM. For AMD, we would like to thank Vamsi Boppana, Negin Oliver, and Phil Guido, along with their respective teams for their invaluable and continuing support. From IBM, we would like to thank Alan Peacock, Jay Jubran, and Brendan Kinkade, along with their respective teams for their incredible work in putting together the cluster at short notice and working closely with us on its operation and governance. From the Zyphra side, we would additionally like to thank Steven Brook for his work negotiating and organizing the agreement and delivery of the cluster, and Paul White for his support developing further our strategic relationships with both parties.

\clearpage

\bibliographystyle{tmlr}
\bibliography{main}

\clearpage



\appendices

\section{Cluster Details}
\label{app:cluster}

\subsection{Compute Nodes}

Each compute node contains:

\begin{itemize}
    \item \textbf{GPUs}: 8 MI300X~\citep{amd2024cdna3} GPUs connected via InfinityFabric intra-node interconnect
    \item \textbf{RAM}: 2\,TB of DDR5 RAM. Specifically, 16x128\,GB DIMMs of Samsung M321RAJA0MB0-CWMNY, running at 5600 MT/s.
    \item \textbf{CPU}: 2 physical sockets of Intel Xeon Platinum 8570, each with 56 physical cores and 2 threads per core. Each socket is connected to 1\,TB of RAM (8 DIMMs)
    \item \textbf{Networking Cards}: Eight Pollara 400~\citep{amd_pollara_2024} network interface cards (NICs), each at 400Gbps. One Pensando DSC 200 GbE cloud NIC for loading data and checkpoints. 
    \item \textbf{Storage}: 25.6\,TB split into 8 physical NVMe drives (Micron MTFDKCC3T2TGQ-1BK1DABDB), each with 3.2\,TB.
\end{itemize}

\subsubsection{Storage Node}

The storage node contains:

\begin{itemize}
    \item \textbf{RAM}: 256\,GB of DDR5 RAM. Specifically, 16x16\,GB DIMMs of Samsung M321R2GA3BB6-CQKET, running at 4800 MT/s.
    \item \textbf{CPU}: 2 physical sockets of Intel(R) Xeon(R) Gold 6426Y, each with 16 physical cores and 2 threads per core. Each socket is connected to 128\,GB of RAM (8 DIMMs).
    \item \textbf{Networking Cards}: One Pensando DSC 100 GbE cloud NIC for loading data and checkpoints.
    \item \textbf{Storage}: 120\,TB configured as a single RAID0 array (`/dev/md127`) from 16 physical NVMe drives \\ (Micron\_7450\_MTFDKCC7T6TFR), each with ~7.6\,TB.
\end{itemize}

Each of the above nodes have separate local drives for the OS.

\subsubsection{Login Node}

The login node is a VM that contains:

\begin{itemize}
\item \textbf{RAM}: 80\,GB system memory, installed as 5×16\,GB DIMMs (virtual/QEMU).
\item \textbf{CPU}: 2 sockets of Intel Xeon (Sapphire Rapids), each with 8 cores and 2 threads per core (Virtualized under KVM)
\item \textbf{Storage}: ~1\,TB total, split into 3 virtual disks: vda (100\,GB), vdb (520\,GB), and vdc (520\,GB)
\end{itemize}

\section{Storage Node Sizing and I/O Calculations}
\label{app:storage-io}

We analyze shared-storage requirements for dataset reads during training, assuming Megatron-style pretokenized corpora accessed via \texttt{mmap} or buffered reads on a dedicated storage fabric. Large sequential checkpoint writes are throughput-bound rather than IOPS-bound and are not addressed here.

Let $G$ be the global batch size, $s$ the sequence length, $b$ bytes per token, $P$ the storage page size, $t$ the iteration time, and $I_{\max}$ the sustainable IOPS capacity. We introduce a \emph{scatter factor} $\sigma \ge 1$ to measure how far our dataset accesses (metadata and index touches, page-cache misses, small random seeks) in reality differ from perfectly contiguous reads. Each iteration reads $G \cdot s \cdot b$ bytes, requiring
\begin{align}
N_{\text{IO/iter}}
\;=\;
\sigma \cdot \left\lceil \frac{G \cdot s \cdot b}{P} \right\rceil
\end{align}
effective I/O operations. The sustained IOPS requirement is therefore
\begin{align}
\text{IOPS}_{\text{needed}}
\;=\;
\frac{\sigma}{t}\cdot \left\lceil \frac{G \cdot s \cdot b}{P} \right\rceil ,
\label{eq:iops-needed}
\end{align}
and the break-even iteration time under budget $I_{\max}$ is
\begin{align}
t_{\text{break}}
\;=\;
\frac{\sigma}{I_{\max}}\cdot \left\lceil \frac{G \cdot s \cdot b}{P} \right\rceil .
\label{eq:t-break}
\end{align}

We can attempt to estimate the Scatter Factor \texorpdfstring{$\sigma$}{σ}.
Let $m$ denote the average count of \emph{additional} page faults per sample (metadata, \texttt{*.idx} probe, doc-boundary straddle, or cold read). If the ideal pages-per-sample are $(s \cdot b)/P$, an effective and practical approximation is
\begin{align}
\sigma
\;\approx\;
1 \;+\; \frac{m \cdot P}{s \cdot b}.
\label{eq:sigma}
\end{align}
This interpolates between highly contiguous, warm-cache access ($\sigma \to 1$) and fragmented, small-document regimes ($\sigma > 1$). In our experience with well-packed Megatron datasets, $\sigma \in [1,2]$ is typical; heavily fragmented or multi-shard random-seek workloads can observe $\sigma \in [2,8]$.

For the ZAYA1 training run with $G{=}4096$, $s{=}4096$, $b{=}4$~B, $P{=}4096$~B, $t{=}2.5$~s, and $I_{\max}{=}70{,}000$ IOPS, each iteration reads 64~MiB across 16,384 pages. This requires approximately $6{,}554 \cdot \sigma$ IOPS with break-even time $t_{\text{break}} \approx 0.234 \cdot \sigma$~s. At the observed $t{=}2.5$~s, the run remains comfortably above $t_{\text{break}}$ even for $\sigma{=}8$, confirming our 70k IOPS storage budget is sufficient.

\section{Compressed Convolutional Attention}
\label{app:cca}

We review the architecture of the Compressed Convolutional Attention block \citep{cca}. 

\begin{figure*}[htbp]
    \centering
    \includegraphics[width=.3\textwidth]{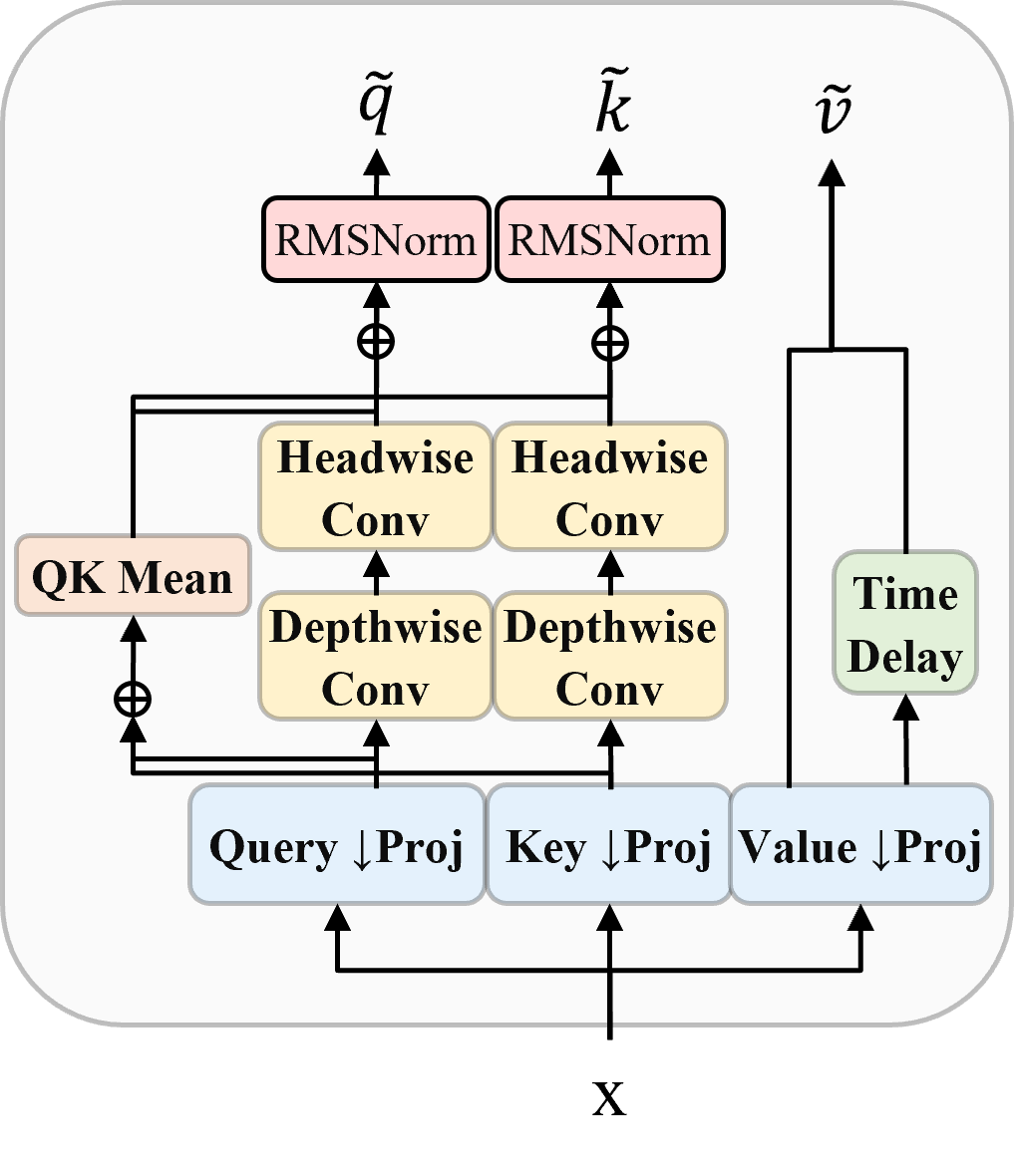}
    \caption*{Compressed Convolutional Attention}
\end{figure*}

CCA is a modification of the attention block which performs attention entirely in a compressed latent space, thus achieving considerable savings in both memory and FLOP costs of attention. CCA outperforms alternative methods such as GQA and MLA in both perplexity and training and inference FLOPs while enabling high rates of KV-cache compression which is crucial for rapid decoding. 

CCA is comprised of several core components:
\begin{itemize}
    \setlength{\itemsep}{0pt} 
    \item \textbf{Low-Rank Projections:} The low-rank down projections for compute and memory decrease.
    \item \textbf{Sequence-Mixing Convolutions:} The short convolution and a grouped head-wise convolution act as a lightweight preconditioner prior to attention.
    \item \textbf{Value Head Time-Delay:} A time delay applied to half of the value heads.
    \item \textbf{Skip Connections and Normalization:} The Query/Key mean skip connections and an RMSnorm layer with a head-wise temperature applied to keys only.
\end{itemize}

We find that these additions -- especially the convolution --  to the standard attention block are vital in providing the expressivity and nonlinearity which can let CCA match and exceed the performance of full attention while requiring substantially less compute and memory. 

The CCA block can also operate in 'GQA-mode' where multiple kv-heads are shared across queries. This provides further savings to decoding due to additional compression of the KV cache. We call this combined method CCGQA. In practice for ZAYA1-base we utilize CCGQA with 2 kv heads for 8 query heads - on top of the $2\times$ compression of the queries - thus achieving a $8\times$ compression of the KV cache vs full multi-head attention. This is perhaps the highest achieved compression of any model of this scale.

\section{Communication Results}
\label{app:communication-results}

While there wasn't sufficient room in the main paper, we include the All-to-All communication operation results in this appendix. Both inter-node and intra-node are included. The expert block of ZAYA1 was able to fit within HBM, but subsequent models will be too large, and expert-parallelism will need to be used. Therefore, we intend to co-design the expert-parallelism implementation of our training framework with the underlying intra-node fabric in order to alleviate the costly All-to-All operation when shuffling tokens to/from experts in every layer.

\begin{figure}[htbp]
    \centering
    \begin{minipage}[b]{0.45\textwidth}
        \centering
        \includegraphics[width=\textwidth]{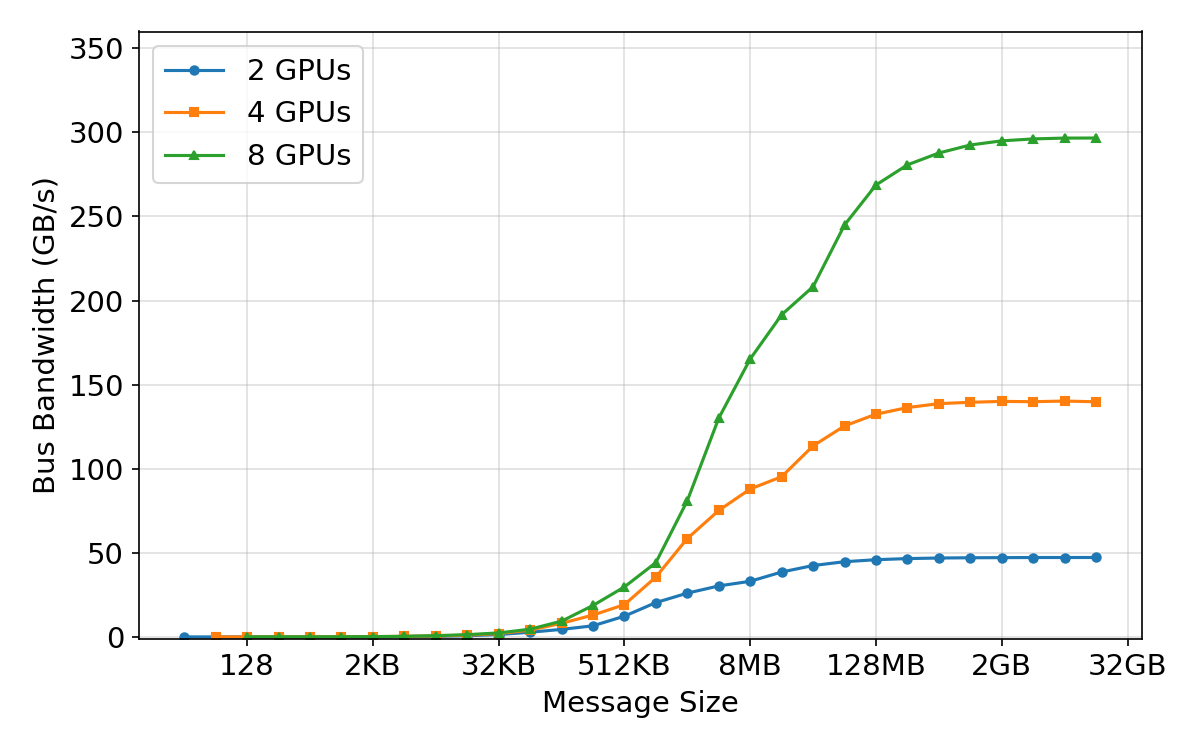}
        \caption*{(a) All-to-All Bus Bandwidth}
    \end{minipage}
    \hfill
    \begin{minipage}[b]{0.45\textwidth}
        \centering
        \includegraphics[width=\textwidth]{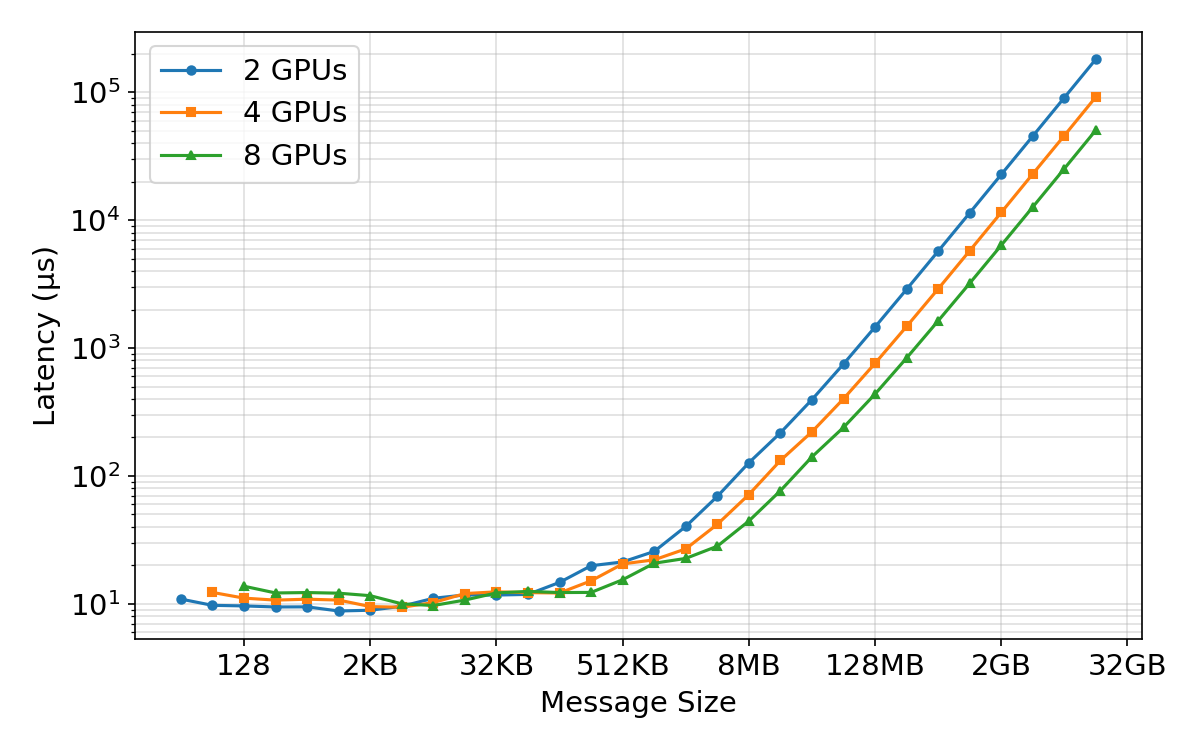}
        \caption*{(b) All-to-All Latency}
    \end{minipage}
    
    \caption{RCCL All-to-All collective operation performance across InfinityFabric within a node, showing bus bandwidth (left) and latency (right).}
    \label{fig:rccl-alltoall-intranode}
\end{figure}

\begin{figure}[htbp]
    \centering
    \begin{minipage}[b]{0.45\textwidth}
        \centering
        \includegraphics[width=\textwidth]{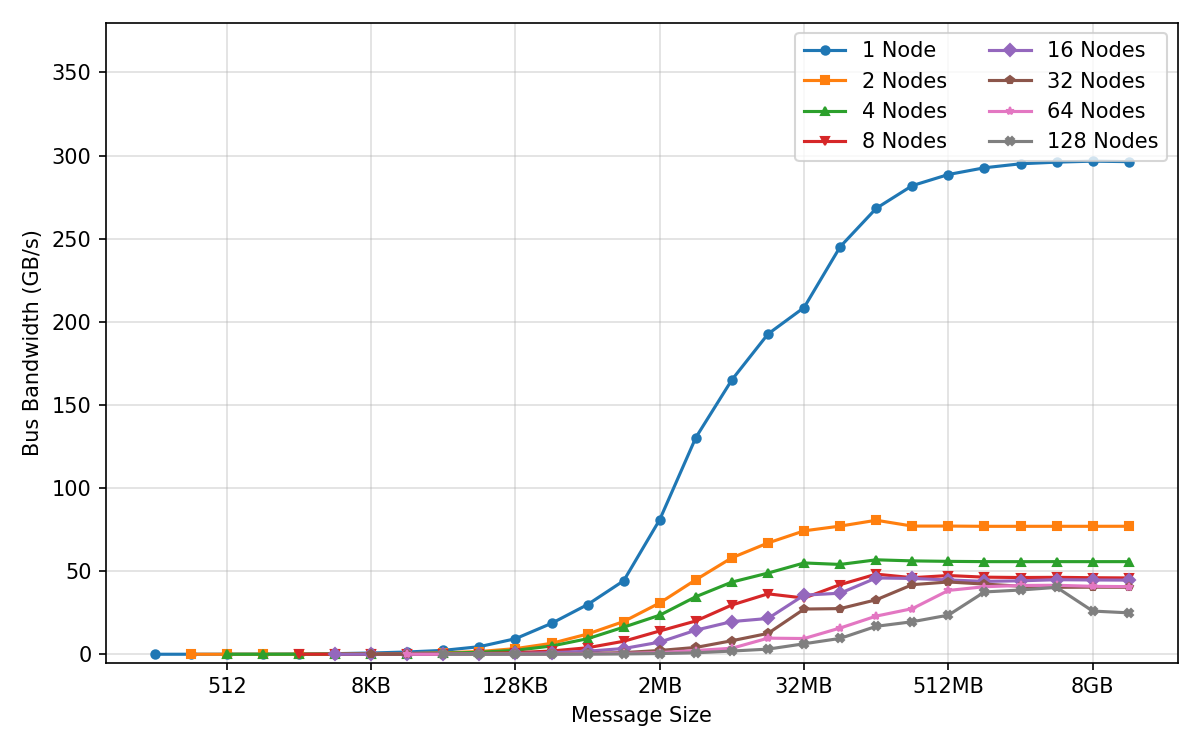}
        \caption*{(a) All-to-All Bus Bandwidth}
    \end{minipage}
    \hfill
    \begin{minipage}[b]{0.45\textwidth}
        \centering
        \includegraphics[width=\textwidth]{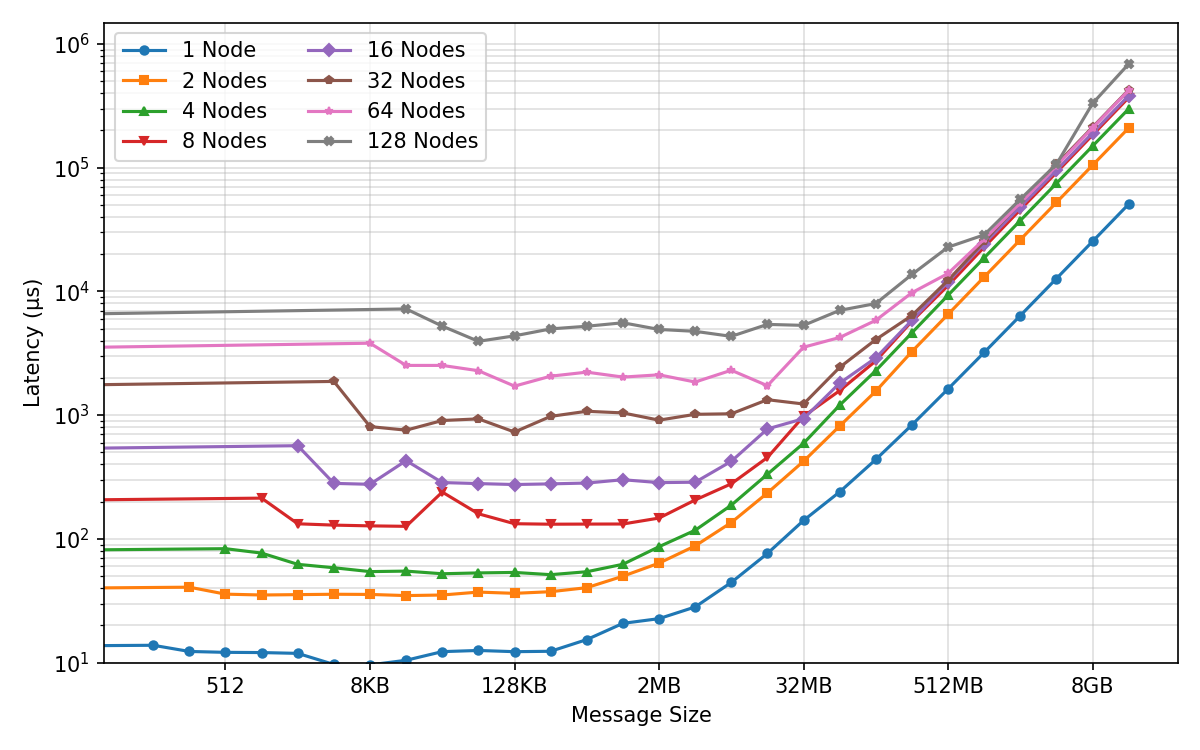}
        \caption*{(b) All-to-All Latency}
    \end{minipage}
    
    \caption{RCCL All-to-All collective operation performance scaling across multiple nodes, showing bus bandwidth (left) and latency (right) with varying node counts.}
    \label{fig:rccl-alltoall-internode}
\end{figure}

\end{document}